\pdfoutput=1

\documentclass[11pt]{article}

\usepackage[]{acl}

\usepackage{times}
\usepackage{latexsym}
\usepackage{times}
\usepackage{float}
\usepackage{latexsym}
\usepackage{multirow}
\usepackage{url}
\usepackage{tabularx}
\usepackage{latexsym}
\usepackage{balance}
\usepackage{bm}
\usepackage{amssymb}
\usepackage{amsmath}
\usepackage{makecell}
\usepackage{multirow}
\usepackage{subfig}
\usepackage{graphicx} 
\usepackage{color}
\usepackage{colortbl}
\usepackage{textcomp}
\usepackage{CJK}
\usepackage{enumitem}
\usepackage{booktabs}
\usepackage{microtype}
\usepackage{arydshln}
\usepackage{amsfonts}
\usepackage{wasysym}
\usepackage{eufrak}
\usepackage{pifont}
\usepackage[normalem]{ulem}
\usepackage[T1]{fontenc}
\usepackage{tcolorbox}
\usepackage{algorithm,algorithmic}
\usepackage{xcolor,soul}
\usepackage{ellipsis}
\tcbuselibrary{breakable}
 \usepackage[normalem]{ulem}
 \useunder{\uline}{\ul}{}

\definecolor{light_blue}{rgb}{0.80,0.85,1.0}
\definecolor{light_green}{rgb}{0.80,1.0,0.85}
\definecolor{light_red}{rgb}{1.0,0.85,0.80}
\definecolor{med_red}{rgb}{1.0,0.55,0.50}
\definecolor{light_orange}{rgb}{1.0,0.72,0.5}

\usepackage[utf8]{inputenc}

\usepackage{microtype}

\DeclareMathOperator*{\argmin}{arg\,min}

\title{KaFT: Knowledge-aware Fine-tuning for Boosting LLMs' Domain-specific Question-Answering Performance}

\author{%
  Qihuang~Zhong$^{1}$,
  Liang~Ding$^{2}$,
  Xiantao~Cai$^{1*}$,
  \textbf{Juhua~Liu}$^{1}$\thanks{~~Corresponding Authors: Xiantao Cai (e-mail: caixiantao@whu.edu.cn), Juhua Liu (e-mail: liujuhua@whu.edu.cn)},
  \textbf{Bo~Du}$^{1}$,
  \textbf{Dacheng~Tao}$^{3}$ \\
  \fontsize{9.0pt}{\baselineskip}\selectfont $^{1}$ School of Computer Science, National Engineering Research Center for Multimedia Software, Institute of Artificial Intelligence \\ 
  \fontsize{9.0pt}{\baselineskip}\selectfont  and Hubei Key Laboratory of Multimedia and Network Communication Engineering, Wuhan University, China \\
  \fontsize{9.0pt}{\baselineskip}\selectfont $^{2}$ The University of Sydney, Australia \quad $^{3}$ Nanyang Technological University, Singapore \\
\fontsize{9.0pt}{\baselineskip}\selectfont \texttt{\{zhongqihuang, caixiantao, liujuhua, dubo\}@whu.edu.cn}\\
\fontsize{9.0pt}{\baselineskip}\selectfont \texttt{\{liangding.liam, dacheng.tao\}@gmail.com}
}

\begin{document}
\maketitle

 \begin{abstract}
Supervised fine-tuning (SFT) is a common approach to improve the domain-specific question-answering (QA) performance of large language models (LLMs). However, recent literature reveals that due to the conflicts between LLMs' internal knowledge and the context knowledge of training data, vanilla SFT using the full QA training set is usually suboptimal. In this paper, we first design a query diversification strategy for robust conflict detection and then conduct a series of experiments to analyze the impact of knowledge conflict. We find that 1) training samples with varied conflicts contribute differently, where SFT on the data with large conflicts leads to catastrophic performance drops; 2) compared to directly filtering out the conflict data, appropriately applying the conflict data would be more beneficial. Motivated by this, we propose a simple-yet-effective Knowledge-aware Fine-tuning (namely KaFT) approach to effectively boost LLMs' performance. The core of KaFT is to adapt the training weight by assigning different rewards for different training samples according to conflict level. Extensive experiments show that KaFT brings consistent and significant improvements across four LLMs. More analyses prove that KaFT effectively improves the model generalization and alleviates the hallucination.
\end{abstract}
 \section{Introduction}
\label{sec_intro}
While large language models (LLMs)~\cite{openai2023gpt4,dubey2024llama,zhao2023survey} have showcased powerful general-purpose capabilities, they often struggle to handle domain-specific question-answering (QA) tasks, \textit{e.g.}, medical QA~\cite{labrak2024biomistral}. Hence, supervised fine-tuning (SFT), aiming to activate LLMs' internal knowledge and align LLMs' output with the desired behavioral norms, is usually required~\cite{zhou2024lima,zhang2024enhancing}. 

However, recent literature~\cite{ren2024learning,gekhman2024does} reveals that domain-specific SFT usually suffers from a crucial problem: \textit{\textbf{knowledge conflict}}, which is the discrepancy between the LLMs' internal knowledge and the context knowledge of training data~\cite{xu2024knowledge}. Due to the long-tail distribution and timeliness of pretraining corpora, LLMs might struggle to learn sufficient domain-specific knowledge during pretraining. Conversely, SFT training datasets usually contain more up-to-date and professional knowledge. Unfortunately, SFT fails to learn additional knowledge~\cite{ren2024learning}, and enforcing LLMs to learn new knowledge through SFT would easily damage their prior abilities and lead to hallucination~\cite{gekhman2024does}. 

\begin{figure}[t]
\centering
\includegraphics[width=0.42\textwidth]{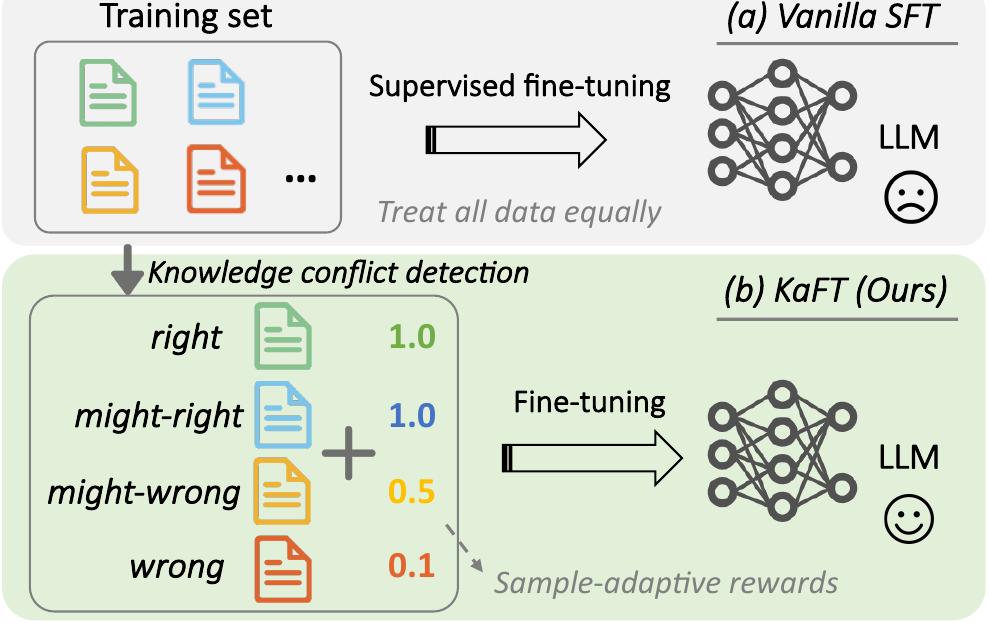}
    \caption{Comparison between \textbf{(a) vanilla SFT} and \textbf{(b) our KaFT}. Different from vanilla SFT treating all training data equally, KaFT uses sample-adaptive rewards to facilitate more effective learning of LLMs.}
    \label{fig:method}
\end{figure}

To tackle this problem, some empirical studies have been conducted~\cite{ren2024learning,gekhman2024does,ye2024empirical}. For instance, \citet{ren2024learning} employ in-context learning (ICL)~\cite{brown2020language} to probe LLMs' internal knowledge and determine whether it conflicts with the training data. Based on this, they analyze the behavior of LLMs after SFT with conflict data. Despite providing some insightful findings, they still have some shortcomings: 1) the proposed conflict detection methods are simply based on ICL, which is sensitive to few-shot examples and might introduce bias into the results~\cite{min2022rethinking,ye2024empirical}; 2) they alleviate the negative effect of knowledge conflict by directly filtering the conflict data, 
while neglecting how to make full use of these data.

To this end, we first improve the ICL-based conflict detection with a \textit{query diversification} strategy to reduce the bias of few-shot examples. Based on it, we conduct a series of preliminary analyses to reveal the impact of knowledge conflict.
Specifically, we calculate the conflict score for each training data and split the training set evenly into four subsets with varied conflicts. By fine-tuning LLMs with different subsets, we find that:
\begin{itemize}
	\item Different subsets contribute differently, where SFT on the individual subset with more conflicts causes catastrophic performance drops. 
	\item Compared to directly filtering the subset with more conflicts, appropriately applying these data might be more beneficial.
\end{itemize}

 Based on these observations, we recognize that \textit{not all training samples are equally important for SFT}, and LLMs should pay different attention to different samples. Motivated by this, we proposed a simple-yet-effective \textbf{K}nowledge-\textbf{a}ware \textbf{F}ine-\textbf{T}uning (namely \textbf{KaFT}) approach to effectively boost LLMs' QA performance. As illustrated in Figure~\ref{fig:method}, the core of KaFT is to assign different rewards to varied subsets and use these rewards to adapt the learning of LLMs.
 Specifically, for the data with more conflicts, KaFT assigns a smaller reward to alleviate its negative effect. Conversely, for the data with fewer conflicts, KaFT uses a larger reward to encourage its learning. By doing so, KaFT can not only avoid overfitting to conflict data, but also effectively activate its internal knowledge for more efficient domain adaptation.

We mainly evaluate our KaFT in the medical QA applications upon four popular LLMs, including LLaMA3-8B/3B~\cite{dubey2024llama}, Qwen1.5-7B~\cite{bai2023qwen}, and Mistral-7B~\cite{jiang2023mistral}. 
Extensive results show that KaFT surpasses the other baselines by a clear margin, and brings consistent and significant performance gains across all LLMs, , \textit{i.e.}, up to \textbf{+5.73\%} and \textbf{+2.40\%} average scores than the base model and vanilla SFT method, respectively. More in-depth analysis prove that KaFT can be expanded to other domain-specific applications. More encouragingly, KaFT improves the model generalization and alleviates the model hallucination problem effectively.

\paragraph{Contributions.} To summarize, our contributions are three-fold: (1) We propose a query diversification strategy for robust conflict detection. Based on it, we conduct a series of preliminary analyses and reveal that training samples with varied conflicts contribute differently.
(2) Motivated by this, we propose a simple-yet-effective knowledge-aware SFT (KaFT) approach, which employs sample-adaptive rewards to boost LLMs' QA performance.
(3) Extensive experiments show that KaFT outperforms the vanilla SFT by a clear margin and improves the model generalization effectively.

 \section{Preliminary}
\label{sec:preliminary}
\subsection{Task Formulation}
Given a domain-specific QA training dataset $D=\{(q_i,o_i,a_i)\}_{i=1}^N$ and a pretrained base LLM $\mathcal{M}_{\theta}$ parameterized by $\theta$, where $q_i$, $o_i$ and $a_i$ denote the question, options and answer, and $N$ denotes the number of all training samples. The goal of SFT is to use the $\mathcal{D}$ to fine-tune $\mathcal{M}_{\theta}$ with supervised learning, \textit{i.e.}, maximum likelihood estimates, and obtain the final adapted LLM $\mathcal{M}_{\theta^*}$.

\begin{figure*}[t]
\centering
    \includegraphics[width=\textwidth]{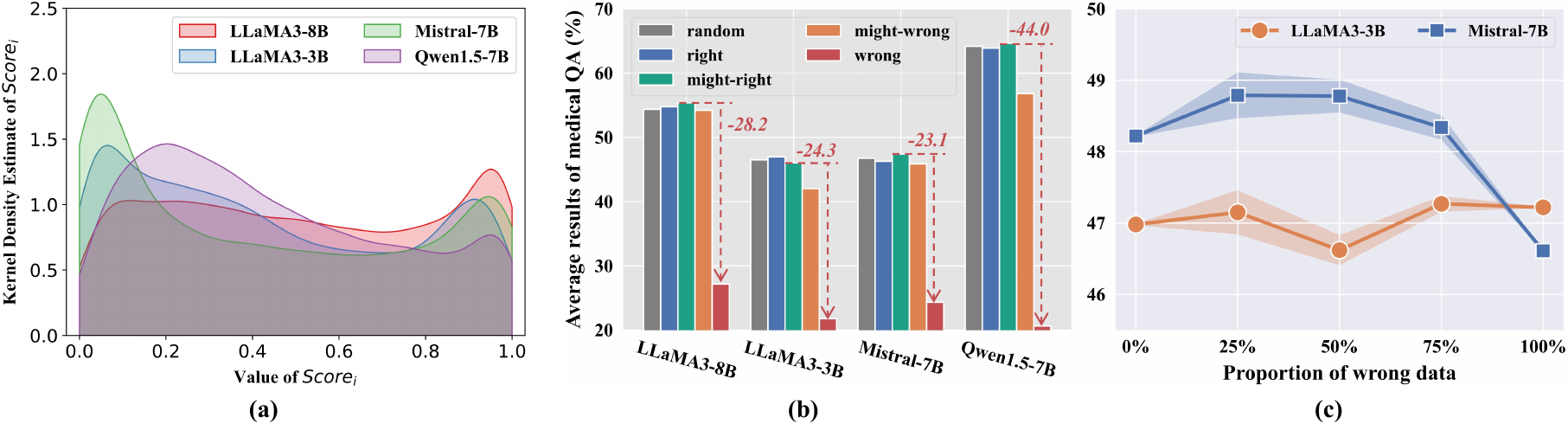}
    \caption{\textbf{(a) Illustration of distributions of $Score_i$ on MedQA across different LLMs}. We use the kernel density estimate for visualizing, where the larger density refers to more training samples. \textbf{(b) Performance comparison (\%) of different subsets}. Note that all subsets hold the same number of training samples. \textbf{(c) Analysis of different proportions of \texttt{wrong} data}. Specifically, we randomly select varied samples from \texttt{wrong} and merge them with the other three subsets. We use three different random seeds for data sampling and report the average results.}
    \label{fig:preliminary_analysis}
\end{figure*}

\subsection{Knowledge Conflict Detection with Query Diversification Strategy}
\label{sec: detect_method}
As mentioned in \S\ref{sec_intro}, SFT usually suffers from the knowledge conflict problem. To detect the knowledge conflicts in $\mathcal{D}$, \citet{ren2024learning} propose an ICL-based probing method. Specifically, they randomly select some training samples as the few-shot examples and utilize them to probe $\mathcal{M}_{\theta}$'s response $r_i$ with greedy decoding, \textit{i.e.}, temperature=0, to each query $(q_i,o_i)$. The response $r_i$ is referred to as the model's parameter knowledge for this question $q_i$. Then, they determine whether the $r_i$ is aligned with reference answer $a_i$, \textit{i.e.}, $\mathbb{I} (r_i=a_i)$, where $\mathbb{I} (\cdot)$ is the indicator function, and regard the misaligned samples as the conflict data. To have a closer look, we provide a case in Appendix~\ref{sec:appendix_prompt}.

Obviously, such a simple ICL-based approach is not robust, as it is sensitive to the few-shot examples and introduce bias. To this end, we improve this method with a \textit{query diversification} strategy. The primary intuition of our strategy is that, if we replace the internal order of options $o_i$ and $\mathcal{M}_{\theta}$ always fails to output the correct answer, $\mathcal{M}_{\theta}$ indeed does not learn the knowledge for the question. In practice, for each data point, we first replace the internal order of $o_i$ and obtain $N_o$ different queries $\{(q_i, o^{j}_i)\}_{j=1}^{N_o}$. Then, we feed the queries into $\mathcal{M}_{\theta}$ to obtain its responses. Moreover, inspired by self-consistency~\cite{wangself}, we set the temperature to 0.7 and sample $N_r$ candidate responses $\{r^j_{i_k}\}_{k=1}^{N_r}$ for each query $(q_i, o^{j}_i)$. Lastly, the knowledge conflict can be measured as:
\begin{equation}
Score_i=\frac{\sum_{j=1}^{N_o}\sum_{k=1}^{N_r}\mathbb{I} (r^j_{i_k}) = a_i)}{N_o \times N_r},
    \label{eq:ka}
\end{equation}
where $Score_i$ denotes the conflict score (larger value refers to fewer conflicts) of $i$-th training data.

\subsection{Empirical Analyses}
\label{sec:preliminary_setting}
\paragraph{Setting.} We use a popular medical QA benchmark, \textit{i.e.}, MedQA~\cite{jin2021disease}, as the testbed, containing 10,178 training data. We perform SFT on four cutting-edge LLMs, including LLaMA3-8B/3B~\cite{dubey2024llama}, Qwen1.5-7B~\cite{bai2023qwen}, and Mistral-7B~\cite{jiang2023mistral}.
The tuned models are evaluated on six medical QA benchmarks, covering the test sets of MedQA,
MedMCQA~\cite{pal2022medmcqa}, MMLU$^*$~\cite{hendrycksmeasuring}\footnote{Following~\citet{singhal2023towards}, we select six medical sub-tasks from MMLU, and denote this subset as MMLU$^*$. Similarly, we also collect the medical sub-tasks from CMMLU~\cite{li2023cmmlu} and denote it as CMMLU$^*$.}), CMB~\cite{wang2024cmb}, CMExam~\cite{liu2024benchmarking}, and CMMLU$^*$~\cite{li2023cmmlu}.
For conflict detection, we set the $N_o$ and $N_r$ to 10.
The distributions of $Score$  are illustrated in Figure~\ref{fig:preliminary_analysis} \textbf{(a)}.

\paragraph{Findings.} To investigate the impact of knowledge conflict, we conduct systematic analyses and empirically observe that:

\begin{table}[]
\resizebox{0.49\textwidth}{!}{%
\begin{tabular}{lcccc}
\toprule
 Train/Test& \texttt{wrong} & \texttt{mig-wr} & \texttt{mig-ri} & \texttt{right} \\ \midrule
\texttt{wrong} &\bf 28.93 & 27.99 & 25.79 & 25.08 \\
\texttt{mig-wr} & 12.58 &\bf 51.57 & 87.42 & 98.75 \\
\texttt{mig-ri} & 5.66 & 48.43 & 88.05 &\bf 99.69 \\
\texttt{right} & 7.55 & 47.17 &\bf 88.36 & 98.75 \\
\bottomrule
\end{tabular}
}
\caption{\textbf{Fine-grained test results of trained LLaMA3-8B models on MedQA}. 
``\texttt{mig-ri}'' and ``\texttt{mig-wr}'' refer to \texttt{might-right} and \texttt{might-wrong} subsets.}
\label{tab:fine_test}
\end{table}

\paragraph{\ding{182} Different subsets contribute differently, where SFT on the individual subset with more conflicts causes catastrophic performance drops.} First, we calculate the $Score$ for each training data and sort the $\mathcal{D}$ based on the score. Then, we split $\mathcal{D}$ evenly into four subsets with varied conflicts, denoted as \texttt{right}, \texttt{might-right}, \texttt{might-wrong} and \texttt{wrong}, where \texttt{right} has less conflicts and \texttt{wrong} has most conflicts. Notably, these subsets have the same number of training samples. We fine-tune the LLMs using different individual subsets and illustrate the comparative results in Figure~\ref{fig:preliminary_analysis} \textbf{(b)}. For reference, we also present the results of SFT on the randomly selected samples. As seen, LLMs tuned with different subsets perform differently. Similar to prior findings~\cite{ren2024learning}, SFT on the \texttt{wrong} leads to catastrophic performance drops, proving the negative effect of knowledge conflict. More interestingly, \texttt{right} is usually not the optimal subset, while \texttt{might-right} performs better among all LLMs. We conjecture that many \texttt{right} samples have been learned by LLMs and struggle to provide useful information. Conversely, \texttt{might-right} can help activate LLMs' internal knowledge and better boost their performance.

Moreover, to further investigate the effect of different subsets, we split the MedQA test set into 4 groups with varied conflicts, and report the fine-grained test results of trained LLaMA3-8B models in Table~\ref{tab:fine_test}. As seen, the model trained with \texttt{wrong} data indeed performs better on the \texttt{wrong} test subset. However, it performs much poorly in the other test subsets, confirming our statement that enforcing the model to learn conflict data would damage its prior abilities.


\paragraph{\ding{183} Compared to directly filtering the subset with more conflicts, appropriately applying these data might be more beneficial.} Intuitively, a straightforward way for alleviating the negative effect of knowledge conflict is to filter out the \texttt{wrong} subset from $\mathcal{D}$ and use the other subsets for SFT. However, as aforementioned, appropriate conflict data might help activate LLMs' internal knowledge and lead to better performance. To verify our conjecture, we introduce different proportions $\lambda$ of conflict data from \texttt{wrong} into the collection of other less-conflict subsets, where $\lambda$ ranges from 0\% to 100\%. The results are illustrated in Figure~\ref{fig:preliminary_analysis} \textbf{(c)}, from which we find that compared to directly filtering the \texttt{wrong} (\textit{i.e.}, $\lambda=0\%$), introducing some conflict data (\textit{e.g.}, $\lambda=25\%$) could be more beneficial. This highlights the necessity of exploring more effective SFT methods to make full use of the conflict data.

 \section{Knowledge-aware Fine-tuning}
\label{sec:method}

Based on the observations in \S\ref{sec:preliminary_setting}, we recognize that \textit{not all training samples are equally important for SFT}, and LLMs should pay different attention to different samples. To this end, we propose a knowledge-aware fine-tuning (KaFT) approach to alleviate the negative effect of knowledge conflict and boost LLMs' performance. In this section, we introduce our KaFT in detail.

\paragraph{Motivation and Intuition.} In addition to the empirical findings in \S\ref{sec:preliminary_setting}, our KaFT is also inspired by a famous cognitive structure migration theory~\cite{ausubel1978educational}, \textit{i.e.}, ``The most important single factor influencing learning is what the student already knows'', which highlights that \textit{paying more attention to the new content relevant to prior learned knowledge can lead to more effective knowledge transfer}. Intuitively, for the data with more conflicts, \textit{e.g.}, \texttt{wrong}, LLMs might easily over-fit the unfamiliar knowledge and lead to poor generalization. In contrast, for data with fewer conflicts, more in-depth learning is beneficial for transferring LLMs' internal knowledge and facilitating effective domain adaptation.

\paragraph{Implementation of KaFT.} In practice, based on our proposed strategy in \S\ref{sec: detect_method}, we first calculate the conflict score $Score_i$ for each training data $(q_i,o_i,a_i)$, and split $\mathcal{D}$ evenly into four subsets with varied conflicts, as done in \S\ref{sec:preliminary_setting}. Then, we assign different rewards for different subsets, where \texttt{might-right} and \texttt{right} hold the larger rewards, and the \texttt{wrong} and \texttt{might-wrong} hold the smaller rewards. Lastly, the rewards are used to control the learning weights of different subsets. The learning objective can be formulated as:
\begin{equation}
\begin{aligned}
  &R_{i}=
    \begin{cases} 
    \alpha, &\text{if} (q_i,o_i,a_i) \in \texttt{wrong}, \\
    \beta, &\text{if} (q_i,o_i,a_i) \in \texttt{might-wrong}, \\ 
    1, &\text{if} (q_i,o_i,a_i) \in \texttt{might-right}, \\ 
    1, &\text{if} (q_i,o_i,a_i) \in \texttt{right},
	\end{cases} \\
  &\theta^{*} := \argmin \mathbb{E}_{(q,o,a,R)\sim \mathcal{D}}[R~\log \mathcal{M}(a|q,o)],
\end{aligned}
\label{eq_kaft}
\end{equation}
where $R_{i}$ denotes the reward for $i$-th training data and $\theta^{*}$ denotes the parameters of final LLM $\mathcal{M}_{\theta^*}$. $\alpha$ and $\beta$ are  rewards between 0 and 1, where $\alpha$ is generally smaller than $\beta$. Empirically, we set $\alpha$ and $\beta$ as 0.1 and 0.5, respectively.

\begin{table*}[ht]
\centering
\setlength{\tabcolsep}{7pt}
\resizebox{\textwidth}{!}{%
\begin{tabular}{clrcllrccc}
\toprule
\multirow{2}{*}{\bf Backbone} & \multicolumn{1}{l}{\multirow{2}{*}{\bf Method}} & \multicolumn{3}{c}{\bf English Medical Benchmark} & \multicolumn{3}{c}{\bf Chinese Medical Benchmark} & \multicolumn{2}{c}{\bf Score} \\ \cmidrule(lr){3-5} \cmidrule(lr){6-8} \cmidrule(lr){9-10}
 & \multicolumn{1}{c}{} & \multicolumn{1}{c}{MedQA} & \multicolumn{1}{l}{MedMCQA} & \multicolumn{1}{c}{MMLU$^*$} & \multicolumn{1}{l}{CMB} & \multicolumn{1}{c}{CMExam} & \multicolumn{1}{c}{CMMLU$^*$} & \multicolumn{1}{l}{Avg.} & \multicolumn{1}{l}{$\Delta \uparrow$ } \\ \midrule
\multirow{5}{*}{Mistral-7B} & Base & 50.98 & 48.31 & 65.07 & 31.73 & 30.68 & 32.90 & 43.28 & - \\
 & Vanilla SFT & \textbf{59.86} & 43.75 & 68.06 & 36.25 & 36.25 & 35.48 & 46.61 & +3.33 \\
 & No-conflict & 58.37 & \textbf{51.11} & \underline{68.69} & \underline{38.09} & \underline{36.92} & \underline{36.17} & \underline{48.22} & \underline{+4.94} \\
 & Self-aligning & 55.62 & \underline{50.99} & \textbf{68.81} & 36.70 & 36.81 & 35.13 & 47.34 & +4.06 \\
&\cellcolor{gray!20} \bf KaFT (Ours) &\cellcolor{gray!20}  \underline{59.54} &\cellcolor{gray!20}  49.87 &\cellcolor{gray!20}  68.47 &\cellcolor{gray!20}  \textbf{38.11} &\cellcolor{gray!20}  \textbf{37.35} &\cellcolor{gray!20}  \textbf{40.73} &\cellcolor{gray!20}  \textbf{49.01} &\cellcolor{gray!20}  \textbf{+5.73} \\ \midrule
\multirow{5}{*}{Qwen1.5-7B} & Base & 48.94 & 50.08 & 62.79 & 74.77 & 76.59 & 70.42 & 63.93 & - \\
 & Vanilla SFT & \underline{52.71} & \underline{50.20} & 61.60 & 74.30 & 76.15 & 70.16 & 64.19 & +0.26 \\
 & No-conflict & 52.24 & \textbf{50.54} & 61.55 & 75.04 & 76.87 & \underline{70.77} & 64.50 & +0.57 \\
 & Self-aligning & 51.77 & 50.11 & \underline{63.18} & \underline{75.23} & \underline{77.05} & 70.71 & \underline{64.67} & \underline{+0.74} \\
 &\cellcolor{gray!20}  \bf KaFT (Ours) &\cellcolor{gray!20}  \textbf{53.57} &\cellcolor{gray!20}  49.82 &\cellcolor{gray!20}  \textbf{63.23} &\cellcolor{gray!20}  \textbf{75.57} &\cellcolor{gray!20}  \textbf{77.27} &\cellcolor{gray!20}  \textbf{72.07} &\cellcolor{gray!20}  \textbf{65.25} &\cellcolor{gray!20}  \textbf{+1.32} \\ \midrule
\multirow{5}{*}{LLaMA3-8B} & Base & 59.62 & 56.51 & 72.66 & 45.50 & 46.04 & 44.62 & 54.16 & - \\
 & Vanilla SFT & 61.82 & 55.75 & 73.34 & 45.99 & 45.85 & 44.95 & 54.62 & +0.46 \\
 & No-conflict & \underline{61.98} & 56.11 & \underline{73.40} & 47.17 & 47.72 & \underline{45.59} & 55.33 & +1.17 \\
 & Self-aligning & 61.35 & \underline{56.56} & 73.00 & \textbf{47.53} & \textbf{48.34} & \textbf{46.55} & \underline{55.55} & +1.39 \\
&\cellcolor{gray!20} \bf KaFT (Ours) &\cellcolor{gray!20}  \textbf{64.10} &\cellcolor{gray!20}  \textbf{56.94} &\cellcolor{gray!20}  \textbf{74.01} &\cellcolor{gray!20}  \underline{47.39} &\cellcolor{gray!20}  \underline{47.89} &\cellcolor{gray!20}  45.44 &\cellcolor{gray!20}  \textbf{55.96} &\cellcolor{gray!20}  \textbf{+1.80} \\ \midrule
\multirow{5}{*}{LLaMA3-3B} & Base & 51.14 & 49.41 & 62.34 & 35.98 & 36.48 & 36.50 & 45.31 & - \\
 & Vanilla SFT & \underline{54.99} & \underline{50.08} & 62.94 & 38.00 & 39.17 & 38.13 & \underline{47.22} & \underline{+1.91} \\
 & No-conflict & 52.95 & 49.20 & \underline{63.76} & \underline{38.91} & \underline{39.78} & 37.27 & 46.98 & +1.67 \\
 & Self-aligning & 52.79 & 49.82 & 62.93 & 38.53 & 38.66 & \underline{38.55} & 46.88 & +1.57 \\
 &\cellcolor{gray!20} \bf KaFT (Ours) &\cellcolor{gray!20}  \textbf{54.52} &\cellcolor{gray!20}  \textbf{50.51} &\cellcolor{gray!20}  \textbf{64.93} &\cellcolor{gray!20}  \textbf{40.19} &\cellcolor{gray!20}  \textbf{39.93} &\cellcolor{gray!20}  \textbf{39.70} &\cellcolor{gray!20}  \textbf{48.30} &\cellcolor{gray!20}  \textbf{+2.99} \\
 \bottomrule
\end{tabular}
}
\caption{\textbf{Performance comparison (\%) on the medical QA benchmarks.} ``Avg.'' denotes the average results, and ``$\Delta \uparrow$'' refers to the gains against the base models. Best results are in \textbf{bold}, and second-best results are \underline{underlined}.}
\label{tab:main}
\end{table*}

\section{Experiments}
\label{sec:experiments}
\subsection{Setup}
\paragraph{Tasks and Datasets.}
Similar to the settings of \S\ref{sec:preliminary_setting}, we mainly apply our KaFT in the medical QA and fine-tune LLMs with the training set of MedQA. The tuned models are evaluated on six in-domain test sets, covering English medical QA (MedQA, MedMCQA, MMLU$^*$) and Chinese medical QA (CMB, CMExam and CMMLU$^*$).
 Moreover, we follow~\citet{ren2024learning} and use the constructed QA test sets from three domains: history, engineering and law, as the out-of-domain (OOD) benchmarks. For evaluation, we use the public \texttt{lm-evaluation-harness} toolkit and report the zero-shot accuracy for each benchmark. The details of all tasks are shown in Appendix~\ref{sec:appendix_tasks}.

\paragraph{Models.}
We conduct extensive experiments on four cutting-edge LLMs across different model architectures and sizes, \textit{i.e.}, LLaMA3-8B/3B~\cite{dubey2024llama}, Qwen1.5-7B~\cite{bai2023qwen}, and Mistral-7B~\cite{jiang2023mistral}. In the implementation of KaFT, the $N_o$ and $N_r$ are set to 10. We train each model with a batch size of 16 and a peak learning rate of 1e-4, except 2e-4 for LLaMA3-3B.  All models are trained with the LoRA~\cite{hulora} for 1 epoch.
 The details of model training and inference can be found in Appendix~\ref{sec:appendix_training}.

\paragraph{Baselines.}
We compare KaFT with a series of baselines: 1) \textit{\textbf{Base}} denotes the original LLMs without SFT, 2) \textit{\textbf{Vanilla SFT}} denotes directly fine-tuning LLMs on the full training set equally, 3) \textit{\textbf{No-conflict}} denotes first removing the conflict data (\texttt{wrong} identified in \S\ref{sec:preliminary_setting}) and then fine-tuning LLMs on the remaining training data, and 4) \textit{\textbf{Self-aligning}}, inspired by~\citet{ren2024learning}, denotes first modifying the answers of \texttt{wrong} to match LLM's internal knowledge (\textit{i.e.}, replacing $a_i$ with $r_i$) and then fine-tuning LLMs on the combination of aligned \texttt{wrong} and the other original subsets.

\begin{table*}[t]
\centering
\resizebox{\textwidth}{!}{%
\begin{tabular}{lccclcccccl}
\toprule
\multicolumn{1}{c}{\multirow{2}{*}{\bf Method}} & \multicolumn{5}{c}{\bf Mistral-7B} & \multicolumn{5}{c}{\bf LLaMA3-3B} \\ \cmidrule(lr){2-6} \cmidrule(lr){7-11}
\multicolumn{1}{c}{} & History & Engineering & Law & Avg. & $\Delta \uparrow$ & History & Engineering & Law & Avg. & $\Delta \uparrow$ \\ \midrule
Base &41.20 & 53.20 & 46.80 & 47.07 & - & 33.60 & 54.80 & 38.40 & 42.27 & -\\
Vanilla SFT &46.00 & 59.20 & \underline{50.00} & 51.73 & +4.66 & 40.00 & 56.40 & 46.40 & 47.60 & +5.33 \\
No-conflict &\underline{45.20} & \textbf{61.20} & 49.60 & \underline{52.00} & +4.93 & \textbf{40.80} & \underline{58.00} & \underline{48.00} & \underline{48.93} & +6.66 \\
Self-aligning &44.40 & 56.80 & 49.20 & 50.13 & +3.06 & 39.60 & 56.40 & \underline{48.00} & 48.00 & +5.73 \\
\rowcolor{gray!20} \bf KaFT (Ours) &\textbf{50.40} & \underline{60.00} & \textbf{51.60} & \textbf{54.00} & \textbf{+6.93} & \underline{40.40} & \textbf{58.40} & \textbf{49.60} & \textbf{49.47} & \textbf{+7.20} \\
\bottomrule
\end{tabular}
}
\caption{\textbf{Performance comparison (\%) of tuned medical LLMs on the out-of-domain QA test sets}. ``Avg.'' denotes the average performance. Best results are in \textbf{bold}, and second-best results are \underline{underlined}.}
\label{tab:main2}
\end{table*}

\subsection{Compared Results}
The main results on medical QA benchmarks and OOD benchmarks are reported in Tables~\ref{tab:main} and~\ref{tab:main2}, respectively. From these results, we can find that:

\paragraph{KaFT surpasses the other baselines by a clear margin.} As seen, compared to the vanilla SFT, ``No-conflict'' usually achieves better performance, highlighting the harmfulness of conflict data. ``Self-aligning'' can sometimes bring further performance gains against ``No-conflict'', \textit{e.g.}, +0.22\% average score in LLaMA3-8B. However, it might lead to worse performance in some cases. One possible reason is that $r_i$ obtained by the method in~\cite{ren2024learning} can not probe LLMs' internal knowledge well, thus leading to some noise.
Conversely, our KaFT surpasses the other counterparts by a clear margin, proving its superiority.

\paragraph{KaFT brings consistent and significant performance gains among all model sizes and types.} We see that KaFT not only achieves remarkable performance for LLaMA3-family models, but is also beneficial to the Qwen and Mistral models. Specifically, compared to the base models, KaFT brings \textbf{+5.73\%}, \textbf{+1.32\%}, \textbf{+1.80\%} and \textbf{+2.99\%} average gains for Mistral-7B, Qwen1.5-7B and LLaMA3-8B/3B, respectively. These results prove the effectiveness and universality of KaFT.

\paragraph{KaFT effectively improves the OOD performance.} Additionally, we evaluate the tuned LLMs on the OOD benchmarks to verify LLMs' robustness. Due to space limitations, we only present the contrastive results of Mistral-7B and LLaMA3-3B models in Table~\ref{tab:main2}. It can be observed that KaFT significantly outperforms the baselines among all domains, indicating that alleviating the negative effect of conflict data can avoid the overfitting of LLMs, continuing to prove KaFT's effectiveness.


\subsection{Ablation Study}
\label{sec:ablation}

\paragraph{Effect of conflict detection methods.}
One of our contributions is to design a query diversification strategy for robust conflict detection. 
Here, to verify its effectiveness, we compare it with some variants: 1) ``-w/o diverse query'' means removing the query diversification process and obtaining multiple responses for the original query. 2) ``-w/o response sampling'' means using greedy decoding to obtain the model responses with the highest probability for diverse queries, respectively. 3) ``-w/o both'' means removing both processes and directly using greedy decoding to obtain the model response for each original query, as done in~\citet{ren2024learning}. After obtaining the responses, we compared them with the references to calculate the conflict score. Based on it, we sort the training data and select the \texttt{wrong} set. Taking the LLaMA3-8B as an example, we present the medical QA results of models tuned with different \texttt{wrong} sets in Table~\ref{tab:ablation1}. As seen, the \texttt{wrong}  selected by our method leads to maximum performance degradation, \textit{i.e.}, our method can effectively detect the conflict data and select the most conflict subset, proving its effectiveness.

\begin{table}[t]
\centering
\setlength{\tabcolsep}{10pt}
\resizebox{0.48\textwidth}{!}{%
\begin{tabular}{lcc}
\toprule
\multicolumn{1}{l}{\bf Method} & \bf Score & $\Delta \downarrow$  \\ \midrule
Random & 54.38 & -  \\ 
\bf Ours &\bf 27.16 &\bf $\downarrow$ 27.22  \\ \hdashline
-w/o diverse query & 38.96 & $\downarrow$ 15.42 \\
-w/o response sampling & 30.15 & $\downarrow$ 24.23  \\
-w/o both & 49.00 & $\downarrow$ 5.38  \\
\bottomrule
\end{tabular}
}
\caption{\textbf{Performance comparison (\%) of \texttt{wrong} sets selected by different conflict detection methods}. The LLaMA3-8B is used as the base model. ``$\Delta \downarrow$'' denotes the performance drops against the random selection, where larger values refer to better performance.}
\label{tab:ablation1}
\end{table}

\begin{figure}[t]
    \centering
    \includegraphics[width=0.43\textwidth]{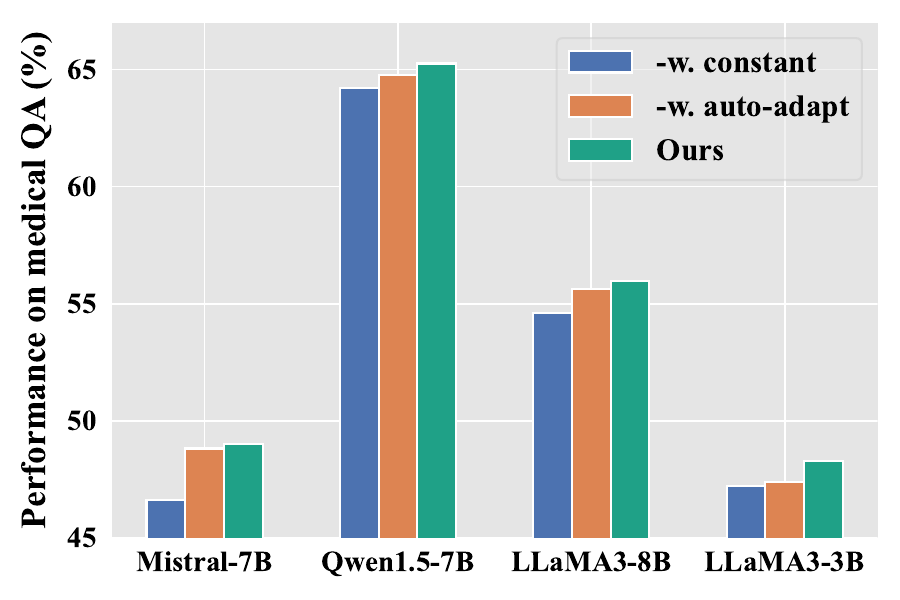}
    \caption{\textbf{Effect of reward strategies in KaFT}. The y-axis denotes the average performance of medical QA.}
    \label{fig:reward}
\end{figure}

\begin{figure}[t]
    \includegraphics[width=0.49\textwidth]{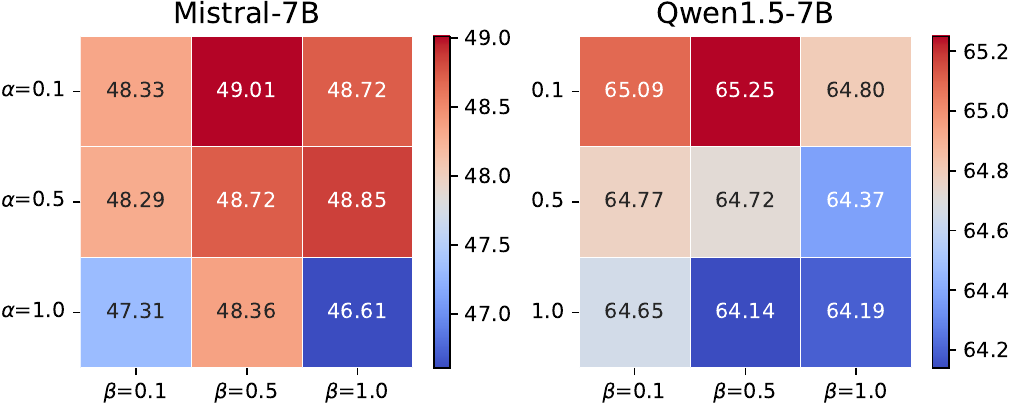}
    \caption{\textbf{Parameter analyses of KaFT}. The y-axis and x-axis denote the varied $\alpha$ and $\beta$, respectively. We report the average results on medical QA benchmarks.}
    \label{fig:reward_analysis}
\end{figure}

\begin{table*}[t]
\centering
\setlength{\tabcolsep}{7pt}
\resizebox{\textwidth}{!}{%
\begin{tabular}{clrcllrccc}
\toprule
\multirow{2}{*}{\bf Backbone} & \multicolumn{1}{l}{\multirow{2}{*}{\bf Method}} & \multicolumn{3}{c}{\bf English Medical Benchmark} & \multicolumn{3}{c}{\bf Chinese Medical Benchmark} & \multicolumn{2}{c}{\bf Score} \\ \cmidrule(lr){3-5} \cmidrule(lr){6-8} \cmidrule(lr){9-10}
 & \multicolumn{1}{c}{} & \multicolumn{1}{c}{MedQA} & \multicolumn{1}{l}{MedMCQA} & \multicolumn{1}{c}{MMLU$^*$} & \multicolumn{1}{l}{CMB} & \multicolumn{1}{c}{CMExam} & \multicolumn{1}{c}{CMMLU$^*$} & \multicolumn{1}{l}{Avg.} & \multicolumn{1}{l}{$\Delta \uparrow$ } \\ \midrule
\multirow{5}{*}{LLaMA3-8B} & Base & 59.62 & 56.51 & 72.66 & 45.50 & 46.04 & 44.62 & 54.16 & - \\
 & KaFT-2 subsets & 61.90 & 56.30 & 73.45 & 48.41 & 48.07 & 46.68 & 55.80 & +1.64 \\
 & KaFT-4 subsets & {64.10} & {56.94} & {74.01} & {47.39} & {47.89} & 45.44 & {55.96} & {+1.80} \\
 & KaFT-8 subsets & 63.45 & 56.63 & 73.82 & 47.93 & 48.01 & 47.12 & 56.16 & +2.00 \\
 \bottomrule
\end{tabular}
}
\caption{\textbf{Analysis of different data partitioning strategies in KaFT}. ``KaFT-* subsets'' denotes that we split the total training dataset into * subsets, and assign different rewards for these subsets.}
\label{tab:appendix_analysis1}
\end{table*}

\begin{table*}[ht]
\centering
\resizebox{\textwidth}{!}{%
\begin{tabular}{lcccccccccc}
\toprule
\multicolumn{1}{c}{\multirow{2}{*}{\bf Method}} & \multicolumn{5}{c}{\bf Mistral-7B} & \multicolumn{5}{c}{\bf LLaMA3-8B} \\ \cmidrule(lr){2-6} \cmidrule(lr){7-11}
\multicolumn{1}{c}{} & QA & Dialogue & Sumarization & Avg. & $\Delta \uparrow$ & QA & Dialogue & Sumarization & Avg. & $\Delta \uparrow$ \\ \midrule
Base & 51.65 & 62.22 & 44.50 & 52.79 & - & \underline{50.79} & 67.27 & \textbf{49.49} & 55.85 & - \\
\texttt{Wrong}-only & 48.92 & 56.71 & 44.40 & 50.01 & \textcolor{red}{-2.78} & 45.55 & 51.73 & 42.56 & 46.61 & \textcolor{red}{-9.24} \\
Vanilla SFT & 50.66 & 69.60 & \textbf{49.24} & 56.50 & \textcolor{green!50!black}{+3.71} & 45.43 & 72.04 & 48.77 & 55.41 & \textcolor{red}{-0.44} \\
No-conflict & \textbf{54.54} & 70.48 & 45.89 & 56.97 & \textcolor{green!50!black}{+4.18} & 49.16 & 71.98 & 48.55 & 56.56 & \textcolor{green!50!black}{+0.71} \\
Self-aligning & \underline{54.22} & \underline{72.41} & 46.11 & \underline{57.58} & \textcolor{green!50!black}{\underline{+4.79}} & 49.67 & \underline{72.24} & 48.64 & \underline{56.85} & \textcolor{green!50!black}{\underline{+1.00}} \\
\rowcolor{gray!20} \bf KaFT (Ours) & 54.20 & \textbf{73.57} & \underline{47.68} & \textbf{58.48} & \textcolor{green!50!black}{\textbf{+5.69}} & \textbf{50.85} & \textbf{72.65} & \underline{48.79} & \textbf{57.43} & \textcolor{green!50!black}{\textbf{+1.58}} \\
\bottomrule
\end{tabular}
}
\caption{\textbf{Performance comparison (\%) on the hallucination evaluation, \textit{i.e.}, HaluEval}~\cite{li2023halueval}. \textcolor{green!50!black}{\textbf{Green}} and \textcolor{red}{\textbf{red}} results refer to the average performance gains and drops against the ``Base'' baseline, respectively. For references, we also report the results of ``\texttt{Wrong}-only'', which fine-tunes LLMs on the individual \texttt{wrong} subset.}
\label{tab:hallucination}
\end{table*}

\paragraph{Effect of reward strategies in KaFT.} As mentioned in \S\ref{sec:method}, KaFT empirically assigns the rewards for subsets with varied conflicts. In this part, we investigate this strategy by comparing it with two variants: 1) ``-w. constant'' refers to the constant reward for all subsets, \textit{i.e.}, $R_i=1.0$, and 2) ``-w. auto-adapt'' refers to using the conflict scores as the rewards, \textit{i.e.}, $R_i=Score_i$. Comparative results of medical QA are illustrated in Figure~\ref{fig:reward}. Both of ours and ``-w. auto-adapt'' outperform the ``-w. constant'' by a clear margin, proving the effectiveness of knowledge-aware SFT. Moreover, ``-w. auto-adapt'' usually performs worse than ours. One possible reason is that it assigns a relatively small reward for the more important \texttt{might-right} subset, thus hindering the activation of LLMs' internal knowledge. Conversely, our strategy can make full use of the training data and achieve the best performance.

\paragraph{Impact of data partitioning strategies in KaFT.}
In our KaFT, we split the total training dataset into 4 subsets and assign different rewards for these subsets. Here, we investigate the effectiveness of KaFT with different data partitioning strategies. Specifically, we split the training data into 2, 4 and 8 subsets based on the conflict scores, respectively. After manually tuning the rewards of the conflict data subset, we train the LLaMA3-8B with our KaFT method and report the comparative results in Table~\ref{tab:appendix_analysis1}. We find that splitting the dataset into more subsets generally results in better performance, as it can achieve more fine-grained and accurate knowledge-aware training. However, more subsets require more manual hyperparameter tuning. Thus, in our work, we split the dataset into 4 subsets for a better trade-off.

\paragraph{Parameter Analysis.} In Eq.~\ref{eq_kaft}, we use two hyperparameters, \textit{i.e.}, $\alpha$ and $\beta$, to control the rewards for \texttt{wrong} and \texttt{might-wrong} subsets. In this study, we analyze their influence by evaluating the performance of KaFT with different $\alpha$ and $\beta$, spanning \{0.1, 0.5, 1.0\}. Figure~\ref{fig:reward_analysis} illustrates the average results of Mistral-7B and Qwen1.5-7B on medical QA benchmarks, from which we find that: 1) Increasing the $\alpha$ leads to a continuous performance decline, confirming the motivation to suppress the learning of conflict data. 2) Increasing the $\beta$ appropriately brings better performance, but too large $\beta$ (\textit{i.e.}, 1.0) is harmful. We conjecture that the \texttt{might-wrong} could contain some conflict data, and overemphasizing its learning would cause over-fitting. More specifically, the case of $\alpha=0.1$ and $\beta=0.5$ performs best, thus leaving as our default experimental settings. 

\section{Discussion}

\begin{table*}[ht]
\resizebox{\textwidth}{!}{%
\begin{tabular}{lcccccccccccc}
\toprule
\multicolumn{1}{c}{\multirow{2}{*}{\bf Method}} & \multicolumn{4}{c}{\bf History $\rightarrow$} & \multicolumn{4}{c}{\bf Engineering $\rightarrow$}  & \multicolumn{4}{c}{\bf Law $\rightarrow$} \\ \cmidrule(lr){2-5} \cmidrule(lr){6-9} \cmidrule(lr){10-13}
\multicolumn{1}{c}{} & History & \multicolumn{1}{l}{Engineering} & Law & Avg. & History & \multicolumn{1}{l}{Engineering} & Law & Avg. & History & \multicolumn{1}{l}{Engineering} & Law & Avg. \\ \midrule
Base & 49.60 & 59.20 & 51.60 & 53.47 & 49.60 & 59.20 & 51.60 & 53.47 & 49.60 & 59.20 & 51.60 & 53.47 \\
Vanilla SFT & 64.40 & 66.00 & 67.20 & 65.87 & 57.60 & 65.20 & \textbf{56.80} & 59.87 & 64.40 & 64.40 & 60.80 & 63.20 \\
No-conflict & 60.00 & \textbf{66.80} & 65.20 & 64.00 & 56.80 & 65.20 & 56.00 & 59.33 & 61.60 & 64.00 & 60.00 & 61.87 \\
Self-aligning & 56.80 & 66.00 & 60.40 & 61.07 & 52.80 & 63.60 & 55.20 & 57.20 & 57.60 & 64.40 & 58.00 & 60.00 \\
\rowcolor{gray!20} \bf KaFT (Ours) & \textbf{66.40} & 66.00 & \textbf{67.60} & \textbf{66.67} & \textbf{58.80} & \textbf{67.20} & \textbf{56.80} & \textbf{60.93} & \textbf{66.00} & \textbf{65.60} & \textbf{61.20} & \textbf{64.27} \\
\bottomrule
\end{tabular}
}
\caption{\textbf{Performance comparison (\%) on more domain-specific QA applications.} Notably, we fine-tune the LLaMA3-8B with the individual domain-specific training set (\textit{i.e.}, History, Engineering, and Law) and evaluate them on all domains' test sets. ``Avg.'' denotes the average performance, and the best results are in \textbf{bold}.}
\label{tab:more_domains}
\end{table*}

\subsection{Does KaFT improve the generalization?}
Intuitively, by alleviating the negative effect of conflict data, KaFT can achieve better model generalization. To verify it, we further analyze its effect from the following aspects:

\begin{figure}[t]
    \centering
    \includegraphics[width=0.45\textwidth]{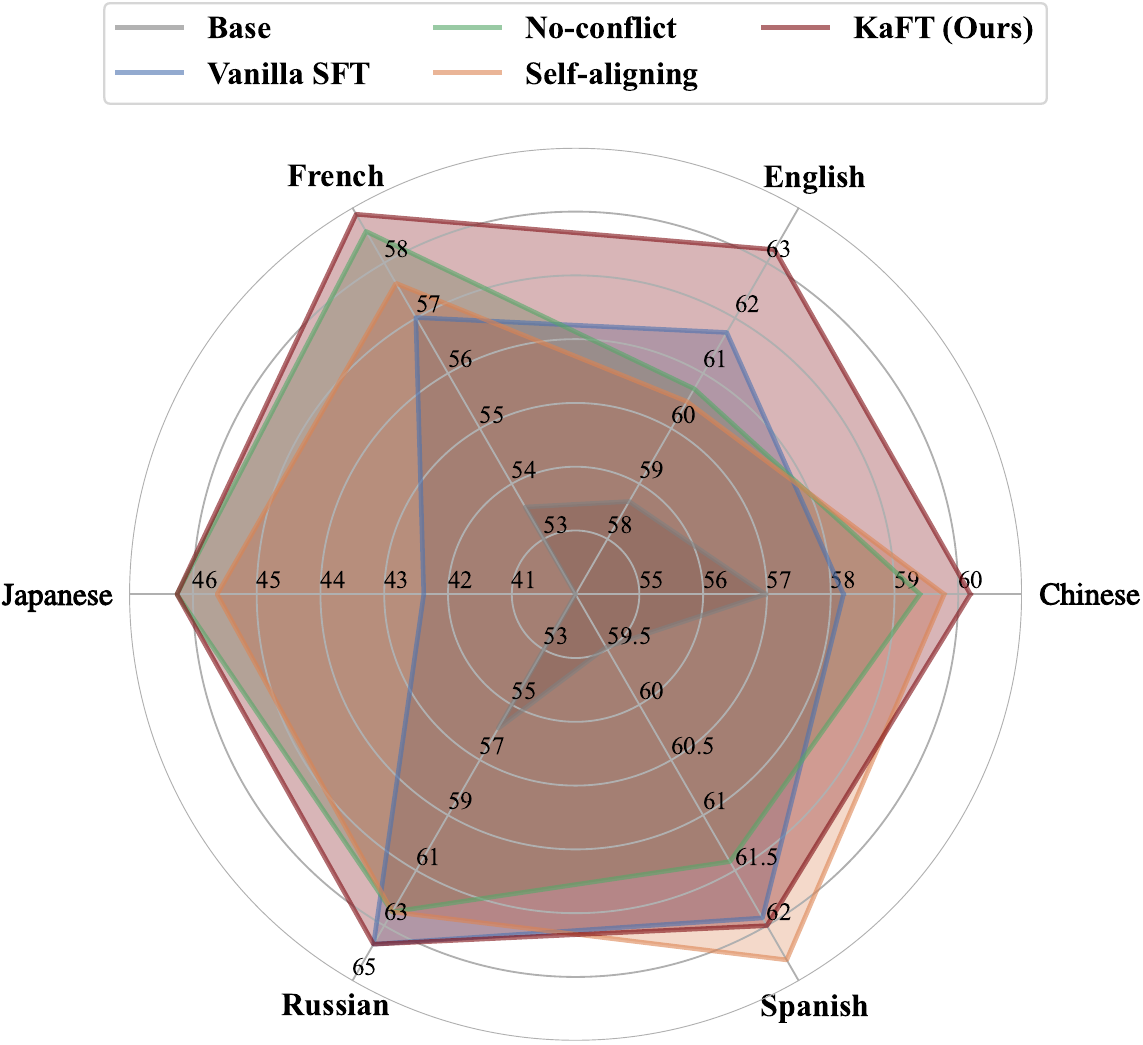}
    \caption{\textbf{Performance comparison (\%) on multilingual medical QA}. LLaMA3-8B is used as base model.}
    \label{fig:mmedbench}
\end{figure}

\paragraph{Multilingual Generalization.} We evaluate the tuned models on a popular multilingual medical QA benchmark, \textit{i.e.}, MMedBench~\cite{qiu2024towards}, consisting of six languages: Chinese, English, French, Japanese, Russian, and Spanish. The comparative results of tuned LLaMA3-8B models are illustrated in Figure~\ref{fig:mmedbench}. As seen, our KaFT brings better performance gains than the other methods across most languages. Specifically, compared to the base model, KaFT achieves \textbf{+4.94\%} average performance gains, especially \textbf{+6.25\%} gains in Japanese and \textbf{+7.81\%} gains in Russian. 

\paragraph{Hallucination Alleviation.}
As stated in the prior work~\cite{gekhman2024does}, fine-tuning with conflict data increases the LLMs' tendency to hallucinate. Here, we investigate this problem by evaluating the tuned models on a popular hallucination detection benchmark, HaluEval~\cite{li2023halueval}. Specifically, the models are required to classify whether a sample contains hallucinated contents from three tasks, \textit{i.e.}, question answering (QA), knowledge-grounded dialogue (Dialogue), and text summarization (Summarization). The results of Mistral-7B and LLaMA3-8B models are reported in Table~\ref{tab:hallucination}. For references, we also report the results of directly fine-tuning on the \texttt{wrong} subset, denoted as ``\texttt{Wrong}-only''. It can be found that enforcing LLMs to learn the new knowledge from conflict data indeed causes serious hallucination, as ``\texttt{Wrong}-only'' and ``Vanilla SFT'' cause up to \textbf{-9.24\%} and \textbf{-0.44\%} average score drops, respectively. More encouragingly, our KaFT can effectively alleviate this side effect and bring up to \textbf{+5.69\%} average score gains against base models. Takeaway: \textbf{\textit{These results prove that our KaFT can indeed bring better model generalization.}}

\subsection{Does KaFT still work in other scenarios?} 


Although our KaFT is mainly evaluated in the medical domain, we believe that it has great potential to expand to more domain-specific applications. To verify it, we conduct additional experiments from three domains: history, engineering, and law. Following~\citet{ren2024learning}, we use the corresponding domain-specific training and test sets, collected from Xiezhi Benchmark~\cite{gu2024xiezhi}. The data statistics are provided in Appendix~~\ref{sec:appendix_tasks}. We fine-tune the LLMs with the individual domain-specific training set and evaluate them on the test sets of all domains. Results of tuned LLaMA3-8B models are reported in Table~\ref{tab:more_domains}, from which we observe that KaFT performs best and brings consistent and significant performance gains among all domains. Takeaway: \textbf{\textit{KaFT not only works well in medical QA, but also can be applied to more domain-specific scenarios}}.
 \section{Related Works}
\label{sec:related}
LLMs~\cite{ouyang2022training,openai2023gpt4,dubey2024llama,liu2024deepseek} have showcased powerful general-purpose capabilities.
However, they might fall short in domain-specific applications, such as medical QA~\cite{labrak2024biomistral}. To this end, many prior works~\cite{singhal2023large,li2023chatdoctor,chen2023meditron,he2025survey} attempt to perform SFT on the domain-specific QA dataset for facilitating domain adaptation. 

Despite achieving remarkable performance,  SFT often faces a critical challenge, \textit{i.e.}, knowledge conflicts. Specifically, since domain-specific SFT is more knowledge-intensive and contains rich professional knowledge that has not been learned during the LLMs' pretraining, there is usually a discrepancy between the LLMs' internal knowledge and the context knowledge of the SFT corpus. More recently, \citet{ren2024learning} reveal that SFT fails to learn additional knowledge and \citet{gekhman2024does} find that enforcing LLMs to learn new knowledge through SFT would easily damage their prior abilities and lead to hallucination. Thus, it is suboptimal to directly fine-tune LLMs using the full SFT training samples equally.

To address this problem, there are few existing works~\cite{ren2024learning,gekhman2024does,ye2024empirical}. However, they still have some shortcomings and struggle to tackle this problem effectively. On the one hand, their conflict detection methods highly rely on ICL~\cite{brown2020language}, which is sensitive to the few-shot examples~\cite{min2022rethinking}. On the other hand, after detecting the conflict data, they mitigate its negative effect by either using early-stopping or filtering out it from the training dataset, while neglecting how to make full use of these conflict data.

 Different from these prior studies, we first design a query diversification strategy to robustly detect the conflict and then propose KaFT to make full use of all training data. The main idea of KaFT is to use sample-adaptive rewards for better guiding the learning of LLMs, 
 which is somewhat similar to prior adaptive-learning methods~\cite{wangopenchat,li2020boosting,kang2020neural,ghorbani2019data,yoon2020data,zhong2024revisiting}. We should emphasize that, although assigning different weights to subsets is a common and intuitive approach, it is non-trivial to determine the weights for different subsets, especially for the domain-specific SFT of LLMs.
Thus, we believe that KaFT is innovative and our work is insightful.
 

 

 \section{Conclusion}
\label{sec:conclusion}
In this paper, we focus on the knowledge conflict problem in the domain-specific SFT, which is critical yet under-explored. Specifically, we propose a query diversification strategy to robustly detect the conflict. Based on it, we conduct a series of preliminary analyses and reveal that different training samples contribute differently, where those with more conflicts would dynamically damage LLMs' abilities. To this end, we further propose a knowledge-aware SFT approach (KaFT). In short, KaFT utilizes sample-adaptive rewards to suppress the negative effect of conflict data and encourage LLMs to activate more relevant knowledge.
Extensive results on medical QA benchmarks demonstrate the effectiveness and universality of KaFT. More encouragingly, in-depth analyses prove that KaFT can achieve better model generalization and alleviate the model hallucination effectively.


\section*{Limitations}
Our work has several potential limitations.
First, given the limited computational budget, we only validate our KaFT on up to 8B LLMs in the main experiments. It will be more convincing if scaling up to super-large model sizes (\textit{e.g.}, 70B) and applying KaFT to more cutting-edge model architectures. 
On the other hand, to better probe LLMs' internal knowledge, we follow the prior studies~\cite{ren2024learning,ye2024empirical} and mainly focus on multiple-choice QA tasks. We will expand our methods to the long-form QA scenarios in future work.

\section*{Ethics Statements}
We take ethical considerations very seriously and strictly adhere to the ACL Ethics Policy. This paper proposes a knowledge-aware fine-tuning framework to improve the domain-specific QA performance of LLMs. It aims to activate LLMs' internal domain-specific knowledge, \textit{e.g.}, medical, instead of encouraging them to learn privacy knowledge that may cause an ethical problem. Moreover, all training and evaluation datasets used in this paper are publicly available and have been widely adopted by researchers. Thus, we believe that this research will not pose ethical issues.


\section*{Acknowledgements}
We are grateful to the anonymous reviewers and the area chair for their insightful comments and suggestions.
This work was supported in part by the National Key Research and Development Program of China under Grant 2023YFC2705700, in part by the National Natural Science Foundation of China under Grant 623B2076, U23B2048 and 62225113, in part by the Innovative Research Group Project of Hubei Province under Grant 2024AFA017, and in part by the Science and Technology Major Project of Hubei Province under Grant 2024BAB046. Dr Tao’s research is partially supported by NTU RSR and Start Up Grants. The numerical calculations in this paper have been done on the supercomputing system in the Supercomputing Center of Wuhan University.

\bibliography{acl2025}

\begin{thebibliography}{39}
\expandafter\ifx\csname natexlab\endcsname\relax\def\natexlab#1{#1}\fi

\bibitem[{Ausubel et~al.(1978)Ausubel, Novak, Hanesian et~al.}]{ausubel1978educational}
David~Paul Ausubel, Joseph~Donald Novak, Helen Hanesian, et~al. 1978.
\newblock Educational psychology: A cognitive view.

\bibitem[{Bai et~al.(2023)Bai, Bai, Chu, Cui, Dang, Deng, Fan, Ge, Han, Huang et~al.}]{bai2023qwen}
Jinze Bai, Shuai Bai, Yunfei Chu, Zeyu Cui, Kai Dang, Xiaodong Deng, Yang Fan, Wenbin Ge, Yu~Han, Fei Huang, et~al. 2023.
\newblock Qwen technical report.
\newblock \emph{arXiv preprint arXiv:2309.16609}.

\bibitem[{Brown et~al.(2020)Brown, Mann, Ryder, Subbiah, Kaplan, Dhariwal, Neelakantan, Shyam, Sastry, Askell et~al.}]{brown2020language}
Tom Brown, Benjamin Mann, Nick Ryder, Melanie Subbiah, Jared~D Kaplan, Prafulla Dhariwal, Arvind Neelakantan, Pranav Shyam, Girish Sastry, Amanda Askell, et~al. 2020.
\newblock Language models are few-shot learners.
\newblock In \emph{Advances in neural information processing systems}.

\bibitem[{Chen et~al.(2023)Chen, Cano, Romanou, Bonnet, Matoba, Salvi, Pagliardini, Fan, K{\"o}pf, Mohtashami et~al.}]{chen2023meditron}
Zeming Chen, Alejandro~Hern{\'a}ndez Cano, Angelika Romanou, Antoine Bonnet, Kyle Matoba, Francesco Salvi, Matteo Pagliardini, Simin Fan, Andreas K{\"o}pf, Amirkeivan Mohtashami, et~al. 2023.
\newblock Meditron-70b: Scaling medical pretraining for large language models.
\newblock \emph{arXiv preprint arXiv:2311.16079}.

\bibitem[{Dubey et~al.(2024)Dubey, Jauhri, Pandey, Kadian, Al-Dahle, Letman, Mathur, Schelten, Yang, Fan et~al.}]{dubey2024llama}
Abhimanyu Dubey, Abhinav Jauhri, Abhinav Pandey, Abhishek Kadian, Ahmad Al-Dahle, Aiesha Letman, Akhil Mathur, Alan Schelten, Amy Yang, Angela Fan, et~al. 2024.
\newblock The llama 3 herd of models.
\newblock \emph{arXiv preprint arXiv:2407.21783}.

\bibitem[{Gekhman et~al.(2024)Gekhman, Yona, Aharoni, Eyal, Feder, Reichart, and Herzig}]{gekhman2024does}
Zorik Gekhman, Gal Yona, Roee Aharoni, Matan Eyal, Amir Feder, Roi Reichart, and Jonathan Herzig. 2024.
\newblock Does fine-tuning llms on new knowledge encourage hallucinations?
\newblock In \emph{Proceedings of the 2024 Conference on Empirical Methods in Natural Language Processing}.

\bibitem[{Ghorbani and Zou(2019)}]{ghorbani2019data}
Amirata Ghorbani and James Zou. 2019.
\newblock Data shapley: Equitable valuation of data for machine learning.
\newblock In \emph{International conference on machine learning}. PMLR.

\bibitem[{Gu et~al.(2024)Gu, Zhu, Ye, Zhang, Wang, Zhu, Jiang, Xiong, Li, Wu et~al.}]{gu2024xiezhi}
Zhouhong Gu, Xiaoxuan Zhu, Haoning Ye, Lin Zhang, Jianchen Wang, Yixin Zhu, Sihang Jiang, Zhuozhi Xiong, Zihan Li, Weijie Wu, et~al. 2024.
\newblock Xiezhi: An ever-updating benchmark for holistic domain knowledge evaluation.
\newblock In \emph{Proceedings of the AAAI Conference on Artificial Intelligence}.

\bibitem[{He et~al.(2025)He, Mao, Lin, Ruan, Lan, Feng, and Cambria}]{he2025survey}
Kai He, Rui Mao, Qika Lin, Yucheng Ruan, Xiang Lan, Mengling Feng, and Erik Cambria. 2025.
\newblock A survey of large language models for healthcare: from data, technology, and applications to accountability and ethics.
\newblock \emph{Information Fusion}.

\bibitem[{Hendrycks et~al.(2020)Hendrycks, Burns, Basart, Zou, Mazeika, Song, and Steinhardt}]{hendrycksmeasuring}
Dan Hendrycks, Collin Burns, Steven Basart, Andy Zou, Mantas Mazeika, Dawn Song, and Jacob Steinhardt. 2020.
\newblock Measuring massive multitask language understanding.
\newblock In \emph{International Conference on Learning Representations}.

\bibitem[{Hu et~al.(2021)Hu, Wallis, Allen-Zhu, Li, Wang, Wang, Chen et~al.}]{hulora}
Edward~J Hu, Phillip Wallis, Zeyuan Allen-Zhu, Yuanzhi Li, Shean Wang, Lu~Wang, Weizhu Chen, et~al. 2021.
\newblock Lora: Low-rank adaptation of large language models.
\newblock In \emph{International Conference on Learning Representations}.

\bibitem[{Jiang et~al.(2023)Jiang, Sablayrolles, Mensch, Bamford, Chaplot, Casas, Bressand, Lengyel, Lample, Saulnier et~al.}]{jiang2023mistral}
Albert~Q Jiang, Alexandre Sablayrolles, Arthur Mensch, Chris Bamford, Devendra~Singh Chaplot, Diego de~las Casas, Florian Bressand, Gianna Lengyel, Guillaume Lample, Lucile Saulnier, et~al. 2023.
\newblock Mistral 7b.
\newblock \emph{arXiv preprint arXiv:2310.06825}.

\bibitem[{Jin et~al.(2021)Jin, Pan, Oufattole, Weng, Fang, and Szolovits}]{jin2021disease}
Di~Jin, Eileen Pan, Nassim Oufattole, Wei-Hung Weng, Hanyi Fang, and Peter Szolovits. 2021.
\newblock What disease does this patient have? a large-scale open domain question answering dataset from medical exams.
\newblock \emph{Applied Sciences}.

\bibitem[{Kang et~al.(2020)Kang, Han, and Hwang}]{kang2020neural}
Minki Kang, Moonsu Han, and Sung~Ju Hwang. 2020.
\newblock Neural mask generator: Learning to generate adaptive word maskings for language model adaptation.
\newblock In \emph{Proceedings of the 2020 Conference on Empirical Methods in Natural Language Processing (EMNLP)}.

\bibitem[{Labrak et~al.(2024)Labrak, Bazoge, Morin, Gourraud, Rouvier, and Dufour}]{labrak2024biomistral}
Yanis Labrak, Adrien Bazoge, Emmanuel Morin, Pierre-Antoine Gourraud, Mickael Rouvier, and Richard Dufour. 2024.
\newblock Biomistral: A collection of open-source pretrained large language models for medical domains.
\newblock In \emph{Findings of the Association for Computational Linguistics: ACL 2024}.

\bibitem[{Li et~al.(2020)Li, Huang, Lan, Feng, Li, and Wang}]{li2020boosting}
Aoxue Li, Weiran Huang, Xu~Lan, Jiashi Feng, Zhenguo Li, and Liwei Wang. 2020.
\newblock Boosting few-shot learning with adaptive margin loss.
\newblock In \emph{Proceedings of the IEEE/CVF conference on computer vision and pattern recognition}.

\bibitem[{Li et~al.(2024)Li, Zhang, Koto, Yang, Zhao, Gong, Duan, and Baldwin}]{li2023cmmlu}
Haonan Li, Yixuan Zhang, Fajri Koto, Yifei Yang, Hai Zhao, Yeyun Gong, Nan Duan, and Timothy Baldwin. 2024.
\newblock Cmmlu: Measuring massive multitask language understanding in chinese.
\newblock In \emph{Findings of the Association for Computational Linguistics: ACL 2024}.

\bibitem[{Li et~al.(2023{\natexlab{a}})Li, Cheng, Zhao, Nie, and Wen}]{li2023halueval}
Junyi Li, Xiaoxue Cheng, Wayne~Xin Zhao, Jian-Yun Nie, and Ji-Rong Wen. 2023{\natexlab{a}}.
\newblock Halueval: A large-scale hallucination evaluation benchmark for large language models.
\newblock In \emph{Proceedings of the 2023 Conference on Empirical Methods in Natural Language Processing}.

\bibitem[{Li et~al.(2023{\natexlab{b}})Li, Li, Zhang, Dan, Jiang, and Zhang}]{li2023chatdoctor}
Yunxiang Li, Zihan Li, Kai Zhang, Ruilong Dan, Steve Jiang, and You Zhang. 2023{\natexlab{b}}.
\newblock Chatdoctor: A medical chat model fine-tuned on a large language model meta-ai (llama) using medical domain knowledge.
\newblock \emph{Cureus}.

\bibitem[{Liu et~al.(2024{\natexlab{a}})Liu, Feng, Xue, Wang, Wu, Lu, Zhao, Deng, Zhang, Ruan et~al.}]{liu2024deepseek}
Aixin Liu, Bei Feng, Bing Xue, Bingxuan Wang, Bochao Wu, Chengda Lu, Chenggang Zhao, Chengqi Deng, Chenyu Zhang, Chong Ruan, et~al. 2024{\natexlab{a}}.
\newblock Deepseek-v3 technical report.
\newblock \emph{arXiv preprint arXiv:2412.19437}.

\bibitem[{Liu et~al.(2024{\natexlab{b}})Liu, Zhou, Hua, Chong, Tian, Liu, Wang, You, Guo, Zhu et~al.}]{liu2024benchmarking}
Junling Liu, Peilin Zhou, Yining Hua, Dading Chong, Zhongyu Tian, Andrew Liu, Helin Wang, Chenyu You, Zhenhua Guo, Lei Zhu, et~al. 2024{\natexlab{b}}.
\newblock Benchmarking large language models on cmexam-a comprehensive chinese medical exam dataset.
\newblock In \emph{Advances in Neural Information Processing Systems}.

\bibitem[{Min et~al.(2022)Min, Lyu, Holtzman, Artetxe, Lewis, Hajishirzi, and Zettlemoyer}]{min2022rethinking}
Sewon Min, Xinxi Lyu, Ari Holtzman, Mikel Artetxe, Mike Lewis, Hannaneh Hajishirzi, and Luke Zettlemoyer. 2022.
\newblock Rethinking the role of demonstrations: What makes in-context learning work?
\newblock In \emph{Proceedings of the 2022 Conference on Empirical Methods in Natural Language Processing}.

\bibitem[{OpenAI(2023)}]{openai2023gpt4}
OpenAI. 2023.
\newblock Gpt-4 technical report.

\bibitem[{Ouyang et~al.(2022)Ouyang, Wu, Jiang, Almeida, Wainwright, Mishkin, Zhang, Agarwal, Slama, Ray et~al.}]{ouyang2022training}
Long Ouyang, Jeffrey Wu, Xu~Jiang, Diogo Almeida, Carroll Wainwright, Pamela Mishkin, Chong Zhang, Sandhini Agarwal, Katarina Slama, Alex Ray, et~al. 2022.
\newblock Training language models to follow instructions with human feedback.
\newblock In \emph{Advances in neural information processing systems}.

\bibitem[{Pal et~al.(2022)Pal, Umapathi, and Sankarasubbu}]{pal2022medmcqa}
Ankit Pal, Logesh~Kumar Umapathi, and Malaikannan Sankarasubbu. 2022.
\newblock Medmcqa: A large-scale multi-subject multi-choice dataset for medical domain question answering.
\newblock In \emph{Conference on health, inference, and learning}.

\bibitem[{Qiu et~al.(2024)Qiu, Wu, Zhang, Lin, Wang, Zhang, Wang, and Xie}]{qiu2024towards}
Pengcheng Qiu, Chaoyi Wu, Xiaoman Zhang, Weixiong Lin, Haicheng Wang, Ya~Zhang, Yanfeng Wang, and Weidi Xie. 2024.
\newblock Towards building multilingual language model for medicine.
\newblock \emph{Nature Communications}.

\bibitem[{Ren et~al.(2024)Ren, Cao, Lin, Liu, Han, Zeng, Wan, Cai, and Sun}]{ren2024learning}
Mengjie Ren, Boxi Cao, Hongyu Lin, Cao Liu, Xianpei Han, Ke~Zeng, Guanglu Wan, Xunliang Cai, and Le~Sun. 2024.
\newblock Learning or self-aligning? rethinking instruction fine-tuning.
\newblock In \emph{Proceedings of the 62nd Annual Meeting of the Association for Computational Linguistics (Volume 1: Long Papers)}.

\bibitem[{Singhal et~al.(2023)Singhal, Azizi, Tu, Mahdavi, Wei, Chung, Scales, Tanwani, Cole-Lewis, Pfohl et~al.}]{singhal2023large}
Karan Singhal, Shekoofeh Azizi, Tao Tu, S~Sara Mahdavi, Jason Wei, Hyung~Won Chung, Nathan Scales, Ajay Tanwani, Heather Cole-Lewis, Stephen Pfohl, et~al. 2023.
\newblock Large language models encode clinical knowledge.
\newblock \emph{Nature}.

\bibitem[{Singhal et~al.(2025)Singhal, Tu, Gottweis, Sayres, Wulczyn, Amin, Hou, Clark, Pfohl, Cole-Lewis et~al.}]{singhal2023towards}
Karan Singhal, Tao Tu, Juraj Gottweis, Rory Sayres, Ellery Wulczyn, Mohamed Amin, Le~Hou, Kevin Clark, Stephen~R Pfohl, Heather Cole-Lewis, et~al. 2025.
\newblock Toward expert-level medical question answering with large language models.
\newblock \emph{Nature Medicine}.

\bibitem[{Wang et~al.(2024{\natexlab{a}})Wang, Cheng, Zhan, Li, Song, and Liu}]{wangopenchat}
Guan Wang, Sijie Cheng, Xianyuan Zhan, Xiangang Li, Sen Song, and Yang Liu. 2024{\natexlab{a}}.
\newblock Openchat: Advancing open-source language models with mixed-quality data.
\newblock In \emph{The Twelfth International Conference on Learning Representations}.

\bibitem[{Wang et~al.(2024{\natexlab{b}})Wang, Chen, Dingjie, Zhiyi, Chen, Xiao, Chen, Jiang, Li, Wan et~al.}]{wang2024cmb}
Xidong Wang, Guiming Chen, Song Dingjie, Zhang Zhiyi, Zhihong Chen, Qingying Xiao, Junying Chen, Feng Jiang, Jianquan Li, Xiang Wan, et~al. 2024{\natexlab{b}}.
\newblock Cmb: A comprehensive medical benchmark in chinese.
\newblock In \emph{Proceedings of the 2024 Conference of the North American Chapter of the Association for Computational Linguistics: Human Language Technologies (Volume 1: Long Papers)}.

\bibitem[{Wang et~al.(2023)Wang, Wei, Schuurmans, Le, Chi, Narang, Chowdhery, and Zhou}]{wangself}
Xuezhi Wang, Jason Wei, Dale Schuurmans, Quoc~V Le, Ed~H Chi, Sharan Narang, Aakanksha Chowdhery, and Denny Zhou. 2023.
\newblock Self-consistency improves chain of thought reasoning in language models.
\newblock In \emph{The Eleventh International Conference on Learning Representations}.

\bibitem[{Xu et~al.(2024)Xu, Qi, Guo, Wang, Wang, Zhang, and Xu}]{xu2024knowledge}
Rongwu Xu, Zehan Qi, Zhijiang Guo, Cunxiang Wang, Hongru Wang, Yue Zhang, and Wei Xu. 2024.
\newblock Knowledge conflicts for llms: A survey.
\newblock In \emph{Proceedings of the 2024 Conference on Empirical Methods in Natural Language Processing}.

\bibitem[{Ye et~al.(2024)Ye, Yang, Zhang, Gui, Huang, Wang, Shi, and Fan}]{ye2024empirical}
Junjie Ye, Yuming Yang, Qi~Zhang, Tao Gui, Xuanjing Huang, Peng Wang, Zhongchao Shi, and Jianping Fan. 2024.
\newblock Empirical insights on fine-tuning large language models for question-answering.
\newblock \emph{arXiv preprint arXiv:2409.15825}.

\bibitem[{Yoon et~al.(2020)Yoon, Arik, and Pfister}]{yoon2020data}
Jinsung Yoon, Sercan Arik, and Tomas Pfister. 2020.
\newblock Data valuation using reinforcement learning.
\newblock In \emph{International Conference on Machine Learning}. PMLR.

\bibitem[{Zhang et~al.(2024)Zhang, Jijo, Setty, Chung, Javid, Vidra, and Clifford}]{zhang2024enhancing}
Liang Zhang, Katherine Jijo, Spurthi Setty, Eden Chung, Fatima Javid, Natan Vidra, and Tommy Clifford. 2024.
\newblock Enhancing large language model performance to answer questions and extract information more accurately.
\newblock \emph{arXiv preprint arXiv:2402.01722}.

\bibitem[{Zhao et~al.(2023)Zhao, Zhou, Li, Tang, Wang, Hou, Min, Zhang, Zhang, Dong et~al.}]{zhao2023survey}
Wayne~Xin Zhao, Kun Zhou, Junyi Li, Tianyi Tang, Xiaolei Wang, Yupeng Hou, Yingqian Min, Beichen Zhang, Junjie Zhang, Zican Dong, et~al. 2023.
\newblock A survey of large language models.
\newblock \emph{arXiv preprint arXiv:2303.18223}.

\bibitem[{Zhong et~al.(2024)Zhong, Ding, Shen, Liu, Du, and Tao}]{zhong2024revisiting}
Qihuang Zhong, Liang Ding, Li~Shen, Juhua Liu, Bo~Du, and Dacheng Tao. 2024.
\newblock Revisiting knowledge distillation for autoregressive language models.
\newblock In \emph{Proceedings of the 62nd Annual Meeting of the Association for Computational Linguistics (Volume 1: Long Papers)}.

\bibitem[{Zhou et~al.(2024)Zhou, Liu, Xu, Iyer, Sun, Mao, Ma, Efrat, Yu, Yu et~al.}]{zhou2024lima}
Chunting Zhou, Pengfei Liu, Puxin Xu, Srinivasan Iyer, Jiao Sun, Yuning Mao, Xuezhe Ma, Avia Efrat, Ping Yu, Lili Yu, et~al. 2024.
\newblock Lima: Less is more for alignment.
\newblock In \emph{Advances in Neural Information Processing Systems}.

\end{thebibliography}

 \appendix
\section{Appendix}
\label{sec:appendix}

\begin{table}[t]
\centering
\resizebox{0.48\textwidth}{!}{%
\begin{tabular}{lll}
\toprule
\textbf{Dataset} & \textbf{\#Training} & \textbf{\#Test} \\
\midrule
\multicolumn{3}{l}{\textit{\textbf{Medical QA}}} \\ \hdashline
MedQA~\cite{jin2021disease} &10,178 & 1,273 \\
MedMCQA~\cite{pal2022medmcqa} &- & 4,183 \\
MMLU$^*$~\cite{hendrycksmeasuring} &- &1,089 \\
\quad - Anatomy &- & 135 \\
\quad - Clinical Knowledge &- & 265 \\
\quad - College Biology &- & 144 \\
\quad - College Medicine &- & 173 \\
\quad - Medical Genetics &- & 100 \\
\quad - Professional Medicine &- & 272 \\ 
\hdashline
CMB~\cite{wang2024cmb}		&- & 9,998 \\
CMExam~\cite{liu2024benchmarking}		&- & 6,607 \\
CMMLU$^*$~\cite{li2023cmmlu} &- &1,140 \\
\quad - Anatomy &- & 148 \\
\quad - Clinical Knowledge &- & 237 \\
\quad - College Medical Statistics &- & 106 \\
\quad - College Medicine &- & 273 \\
\quad - Professional Medicine &- & 376 \\ 
\midrule
\multicolumn{3}{l}{\textit{\textbf{Other domain-specific QA}}} \\ \hdashline
History~\cite{gu2024xiezhi} &8,605 & 250 \\
Engineering~\cite{gu2024xiezhi} &4,805 & 250 \\
Law~\cite{gu2024xiezhi} &6,510 & 250 \\
\midrule
\multicolumn{3}{l}{\textit{\textbf{More in-depth analyses}}} \\ \hdashline
MMedBench~\cite{qiu2024towards} &- &8,178  \\
\quad - Chinese &- & 3,426 \\
\quad - English &- & 1,273 \\
\quad - French &- & 321 \\
\quad - Japanese &- &160  \\
\quad - Russian &- &256  \\
\quad - Spanish &- & 2,742 \\  \hdashline
HaluEval~\cite{li2023halueval} &- &30,000  \\
\quad - question answering &- & 10,000 \\
\quad - knowledge-grounded dialogue &- & 10,000 \\
\quad - text summarization &- & 10,000 \\
\bottomrule
\end{tabular}
}
\caption{\textbf{Statistic information} of all used datasets in our study. ``\#Training'' and ``\#Test'' denote the number of training and test samples, respectively.}
\label{tab:data}
\end{table}

\subsection{Details of Tasks and Datasets}
\label{sec:appendix_tasks}
In this work, to investigate the effectiveness and universality of our KaFT, we conduct extensive experiments on four domain-specific QA applications, covering medical, history, and law. In addition, the multilingual medical QA tasks and hallucination detection tasks are used to reveal the underlying mechanism of our method. Here, we introduce the descriptions of these tasks and datasets in detail. First, we present the statistics of all datasets in Table~\ref{tab:data}. Then, each task is described as:

\paragraph{MedQA.} MedQA~\cite{jin2021disease} consists of questions and corresponding 4-option or 5-option answers in the style of the US Medical License Exam (USMLE). Since it consists of diverse medical knowledge, MedQA is a challenging benchmark and is thus used as our training corpus. Specifically, the training set consists of 10,178 samples, and the test set has 1273 questions.  

\paragraph{MedMCQA.} MedMCQA~\cite{pal2022medmcqa} consists of 4-option multiple-choice QA samples from the Indian medical entrance examinations (AIIMS/NEET). This dataset covers 2.4K healthcare topics and 21 medical subjects. We use the validation set with 4,183 questions for evaluation.

\paragraph{MMLU$^*$.} MMLU~\cite{hendrycksmeasuring} is a comprehensive benchmark, including exam questions from 57 subjects (\textit{e.g.}, STEM and social sciences). Each MMLU subject contains 4-option multiple-choice QA samples. Similar to prior works~\cite{singhal2023towards}, we select 6 subjects that are most relevant to medical and clinical knowledge: Anatomy, Clinical Knowledge, College Biology, College Medicine, Medical Genetics and Professional Medicine. For convenience, we denote this subset as MMLU$^*$.

\paragraph{CMB.} CMB~\cite{wang2024cmb} is a comprehensive medical benchmark in Chinese, designed and rooted entirely within the native Chinese linguistic and cultural framework. It consists of two parts: CMB-Exam, featuring multiple-choice questions from qualification exams, and CMB-Clin, including complex clinical diagnostic questions derived from real case studies. In our experiments, we evaluate the models on the samples with single answers from the test set of CMB-Exam. 

\paragraph{CMExam.} CMExam~\cite{liu2024benchmarking} is sourced from authentic medical licensing exams, containing more than 60K questions. It can reflect the comprehensive coverage of medical knowledge and reasoning required in clinical practice, covering Traditional Medicine Disease Patterns, Digestive System Diseases, Certain Infectious, etc. For evaluation, we select the data with single-choice answers from the test set.

\paragraph{CMMLU$^*$.} CMMLU~\cite{li2023cmmlu} is a comprehensive Chinese benchmark that covers various subjects, including natural sciences, social sciences, engineering, and the humanities. Similar to MMLU-Medical, we also select the subjects that are most relevant to medical and clinical knowledge as the medical QA benchmarks, covering Anatomy, Clinical Knowledge, College Medical Statistics, College Medicine, and Professional Medicine. For convenience, we refer to this subset as CMMLU$^*$ in the main experiments.

\paragraph{Other domain-specific QA.} In addition to the medical QA, we also evaluate our method in the other domains, covering history, engineering, and law. Specifically, we follow~\citet{ren2024learning} and procure the relevant items from the Xiezhi~\cite{gu2024xiezhi} Benchmark for each domain. Xiezhi contains 249587 questions with 516 disciplines, ranging from 13 different categories. Since \citet{ren2024learning} have publicly released the collected dataset, we directly reuse the corresponding training and test sets in our experiments. 

\paragraph{MMedBench.} MMedBench~\cite{qiu2024towards} is a multilingual medical multiple-choice QA benchmark across six primary languages: English, Chinese, Japanese, French, Russian, and Spanish. The entire test set of MMedBench comprises 8,518 QA pairs. For a unified evaluation, we remove the samples with multiple answers and use the filtered 8,178 samples as the evaluation set.

\paragraph{HaluEval.} HaluEval~\cite{li2023halueval} is a large collection of generated and human-annotated hallucinated samples for evaluating the performance of LLMs in recognizing hallucination. It includes 5,000 general user queries with ChatGPT responses and 30,000 task-specific examples from three tasks, \textit{i.e.}, question answering, knowledge-grounded dialogue, and text summarization. In the evaluation, it randomly samples a ground-truth or a hallucinated output for each data. If the text is a hallucinated answer, the LLM should recognize the hallucination and output ``Yes'', which means the text contains hallucinations. If the text is a ground-truth answer, the LLM should output ``No'' indicating that there is no hallucination. The accuracy can evaluate the hallucination, where a larger value means less hallucination. In our study, we use task-specific examples from HaluEval for hallucination evaluation.

\subsection{Training and Evaluation Details}
\label{sec:appendix_training}
For model training, we fine-tune each LLM with a batch size of 16 and a peak learning rate of 1e-4, except 2e-4 for LLaMA3-3B. The warm-up ratio is 0.1 and the maximum tokenizer length is 2,048. All models are trained with LoRA~\cite{hulora} method for 1 epoch. We conduct all experiments on 8 NVIDIA A100 (40GB) GPUs. For conflict detection in KaFT, we set the temperature to 0.7 and sample 10 responses for each query. During evaluation, we set the temperature to 0 for reproducibility. Specifically, we use the widely-used \texttt{lm-evaluation-harness}\footnote{https://github.com/EleutherAI/lm-evaluation-harness} toolkit to measure the zero-shot accuracy of LLMs on multiple-choice QA benchmarks.


\subsection{Prompt Details}
\label{sec:appendix_prompt}
As mentioned in \S\ref{sec:preliminary}, we use the ICL-based method to probe the LLMs' internal domain-specific knowledge for each query. Specifically, we randomly select three samples from the training set as the few-shot examples, and use them to guide the output format of LLMs. Taking the medical QA as an example, we present a case as follows:
\begin{tcolorbox}
[colback=lightgray!20,colframe=darkgray!80,title= Probing for LLMs' internal knowledge]
\label{tab:probing_prompt}
For the following medical question, select one correct answer from A to D.
\newline
\textbf{Question}: A 3900-g (8.6-lb) male infant is delivered at 39 weeks' gestation via spontaneous vaginal delivery. Pregnancy and delivery were uncomplicated but a prenatal ultrasound at 20 weeks showed a defect in the pleuroperitoneal membrane. Further evaluation of this patient is most likely to show which of the following findings?
\newline
\textbf{Options}:
\newline
A. Gastric fundus in the thorax
\newline
B. Pancreatic ring around the duodenum
\newline
C. Hypertrophy of the gastric pylorus
\newline
D. Large bowel in the inguinal canal
\newline
\textbf{Answer}: A
\newline

$\ldots$ (the other two examples)
\newline

For the following medical question, select one correct answer from A to D.
\newline
\textbf{Question}: \texttt{<question>}
\newline
\textbf{Options}:
\newline
A. \texttt{<option\_a>}
\newline
B. \texttt{<option\_b>}
\newline
C. \texttt{<option\_c>}
\newline
D. \texttt{<option\_d>}
\newline
\textbf{Answer}: \texttt{[output]}

\end{tcolorbox}
\noindent where \texttt{<question>} and \texttt{<option>} denote the input question and answer options, \texttt{[output]} denotes the corresponding model response.




\begin{table*}[t]
\centering
\setlength{\tabcolsep}{10pt}
\resizebox{\textwidth}{!}{%
\begin{tabular}{clrcllrccc}
\toprule
\multirow{2}{*}{\bf Backbone} & \multicolumn{1}{l}{\multirow{2}{*}{\bf Subset}} & \multicolumn{3}{c}{\bf English Medical Benchmark} & \multicolumn{3}{c}{\bf Chinese Medical Benchmark} & \multirow{2}{*}{\bf Avg.} \\ \cmidrule(lr){3-5} \cmidrule(lr){6-8}
 & \multicolumn{1}{c}{} & \multicolumn{1}{l}{MedQA} & \multicolumn{1}{l}{MedMCQA} & \multicolumn{1}{l}{MMLU$^*$} & \multicolumn{1}{l}{CMB} & \multicolumn{1}{l}{CMExam} & \multicolumn{1}{l}{CMMLU$^*$} &  \\ \midrule
 \multirow{5}{*}{Mistral-7B} & Random & 55.85 & 50.32 & 66.67 & 36.24 & 36.10 & 35.30 & 46.75 \\
 & \texttt{right} & 53.10 & 47.72 & 66.55 & 37.48 & 36.69 & 36.18 & 46.29 \\
 & \texttt{might-right} & 57.11 & 50.42 & 67.69 & 36.85 & 35.92 & 36.45 & 47.41 \\
 & \texttt{might-wrong} & 56.56 & 50.25 & 64.71 & 35.09 & 34.43 & 34.29 & 45.89 \\
 & \texttt{wrong} & 27.89 & 32.27 & 22.58 & 18.47 & 19.95 & 24.95 & 24.35 \\ \midrule
\multirow{5}{*}{Qwen1.5-7B} & Random & 50.82 & 50.35 & 62.06 & 75.05 & 76.62 & 70.27 & 64.19 \\
 & \texttt{right} & 50.04 & 49.63 & 60.78 & 74.95 & 77.02 & 70.95 & 63.89 \\
 & \texttt{might-right} & 51.37 & 50.35 & 62.87 & 75.61 & 77.37 & 70.32 & 64.65 \\
 & \texttt{might-wrong} & 45.48 & 45.04 & 46.32 & 68.41 & 70.70 & 65.08 & 56.84 \\
 & \texttt{wrong} & 15.63 & 29.69 & 14.72 & 22.79 & 20.13 & 20.87 & 20.64 \\
 \midrule
\multirow{5}{*}{LLaMA3-8B} & Random & 60.80 & 55.42 & 71.68 & 46.84 & 47.15 & 44.42 & 54.38 \\
 & \texttt{right} & 59.31 & 56.59 & 72.20 & 47.37 & 47.54 & 45.64 & 54.77 \\
 & \texttt{might-right} & 60.49 & 55.58 & 73.15 & 48.13 & 48.01 & 46.96 & 55.39 \\
 & \texttt{might-wrong} & 61.67 & 55.49 & 72.58 & 45.76 & 46.12 & 43.75 & 54.23 \\
 & \texttt{wrong} & 23.72 & 30.43 & 37.96 & 23.65 & 22.16 & 25.04 & 27.16 \\ \midrule
\multirow{5}{*}{LLaMA3-3B} & Random & 53.10 & 49.03 & 62.32 & 38.26 & 38.40 & 37.93 & 46.51 \\
 & \texttt{right} & 51.30 & 50.08 & 62.07 & 40.39 & 40.29 & 37.73 & 46.98 \\
 & \texttt{might-right} & 54.20 & 48.29 & 61.09 & 37.09 & 38.14 & 37.48 & 46.05 \\
 & \texttt{might-wrong} & 50.75 & 46.14 & 61.11 & 30.26 & 30.92 & 32.95 & 42.02 \\
 & \texttt{wrong} & 19.25 & 25.32 & 19.80 & 21.02 & 20.63 & 24.58 & 21.77 \\
 \bottomrule
\end{tabular}
}
\caption{\textbf{Full results of Figure~\ref{fig:preliminary_analysis} (b), \textit{i.e.}, comparison of different subsets}. For reference, we also present the results of SFT on the randomly selected samples. Note that all subsets hold the same number of training samples.}
\label{tab:subset_all}
\end{table*}

\begin{table*}[t]
\centering
\setlength{\tabcolsep}{10pt}
\resizebox{\textwidth}{!}{%
\begin{tabular}{clrcllrccc}
\toprule
\multirow{2}{*}{\bf Backbone} & \multicolumn{1}{l}{\multirow{2}{*}{\bf Ratio}} & \multicolumn{3}{c}{\bf English Medical Benchmark} & \multicolumn{3}{c}{\bf Chinese Medical Benchmark} & \multirow{2}{*}{\bf Avg.} \\ \cmidrule(lr){3-5} \cmidrule(lr){6-8}
 & \multicolumn{1}{c}{} & \multicolumn{1}{l}{MedQA} & \multicolumn{1}{l}{MedMCQA} & \multicolumn{1}{l}{MMLU$^*$} & \multicolumn{1}{l}{CMB} & \multicolumn{1}{l}{CMExam} & \multicolumn{1}{l}{CMMLU$^*$} &  \\ \midrule
\multirow{5}{*}{Mistral-7B} & 0\% & 52.95 & 49.20 & 63.76 & 38.91 & 39.78 & 37.27 & 46.98 \\
 & 25\% & 53.73 & 49.15 & 64.52 & 38.86 & 39.14 & 37.47 & 47.15 \\
 & 50\% & 54.91 & 49.94 & 62.04 & 37.81 & 38.70 & 36.32 & 46.62 \\
 & 75\% & 55.85 & 49.32 & 62.93 & 38.17 & 38.99 & 38.37 & 47.27 \\
 & 100\% & 54.99 & 50.08 & 62.94 & 38.00 & 39.17 & 38.13 & 47.22 \\
  \midrule
 \multirow{5}{*}{LLaMA3-3B} & 0\% & 58.37 & 51.11 & 68.69 & 38.09 & 36.92 & 36.17 & 48.22 \\
 & 25\% & 58.29 & 51.23 & 69.25 & 37.54 & 37.17 & 39.25 & 48.79 \\
 & 50\% & 60.02 & 50.32 & 69.69 & 36.75 & 36.73 & 39.15 & 48.78 \\
 & 75\% & 61.19 & 51.66 & 68.54 & 35.87 & 35.51 & 37.29 & 48.34 \\
 & 100\% & 59.86 & 43.75 & 68.06 & 36.25 & 36.25 & 35.48 & 46.61 \\
 \bottomrule
\end{tabular}
}
\caption{\textbf{Full results of Figure~\ref{fig:preliminary_analysis} (c), \textit{i.e.}, analysis of ratio of \texttt{wrong} data}. Notably, we randomly select varied samples from the \texttt{wrong} subset and merge them with the other three subsets. We set three different random seeds for data sampling and report the average results in this table.}
\label{tab:wrong_all}
\end{table*}

\begin{table*}[t]
\centering
\setlength{\tabcolsep}{7pt}
\resizebox{\textwidth}{!}{%
\begin{tabular}{clrcllrccc}
\toprule
\multirow{2}{*}{\bf Backbone} & \multicolumn{1}{l}{\multirow{2}{*}{\bf Method}} & \multicolumn{3}{c}{\bf English Medical Benchmark} & \multicolumn{3}{c}{\bf Chinese Medical Benchmark} & \multirow{2}{*}{\bf Avg.} \\ \cmidrule(lr){3-5} \cmidrule(lr){6-8}
 & \multicolumn{1}{c}{} & \multicolumn{1}{l}{MedQA} & \multicolumn{1}{l}{MedMCQA} & \multicolumn{1}{l}{MMLU$^*$} & \multicolumn{1}{l}{CMB} & \multicolumn{1}{l}{CMExam} & \multicolumn{1}{l}{CMMLU$^*$} &  \\ \midrule
 \multirow{5}{*}{LLaMA3-8B} & Random & 60.80 & 55.42 & 71.68 & 46.84 & 47.15 & 44.42 & 54.38 \\ \cmidrule{2-9}
 & Ours & 23.72 & 30.43 & 37.96 & 23.65 & 22.16 & 25.04 & 27.16 \\ 
 & -w/o diverse query & 41.32 & 40.04 & 61.80 & 29.93 & 29.60 & 31.05 & 38.96 \\
 & -w/o response sampling & 29.38 & 26.63 & 45.03 & 27.46 & 27.27 & 25.16 & 30.15 \\
  & -w/o both & 55.22 & 52.88 & 69.61 & 38.10 & 39.14 & 39.07 & 49.00 \\
\bottomrule
\end{tabular}
}
\caption{\textbf{Full results of Table~\ref{tab:ablation1}, \textit{i.e.}, ablation of our conflict detection method}. LLaMA3-8B is used as the base model. Notably, we use different conflict detection to select the \texttt{wrong} subset for training. The worse results mean that the method can detect the conflict data more accurately, \textit{i.e.}, worse results refer to better performance.}
\label{tab:ablation_full}
\end{table*}

\subsection{Full Results}
\label{sec:appendix_full}
Here, we report the full results of experiments in our main paper. Specifically, Table~\ref{tab:subset_all} shows the detailed results of different subsets. Table~\ref{tab:wrong_all} shows the detailed results of varied \texttt{wrong} data. Table~\ref{tab:ablation_full} and Table~\ref{tab:reward_all} show the ablation study of our proposed conflict detection method and KaFT method, respectively. Table~\ref{tab:parameter_analysis_all} shows the detailed results of parameter analyses of $\alpha$ and $\beta$. Table~\ref{tab:mmedbench_all} shows the detailed results on the MMedBench. Please refer to the tables for more details.

\begin{table*}[t]
\centering
\setlength{\tabcolsep}{7pt}
\resizebox{\textwidth}{!}{%
\begin{tabular}{clrcllrccc}
\toprule
\multirow{2}{*}{\bf Backbone} & \multicolumn{1}{l}{\multirow{2}{*}{\bf Method}} & \multicolumn{3}{c}{\bf English Medical Benchmark} & \multicolumn{3}{c}{\bf Chinese Medical Benchmark} & \multirow{2}{*}{\bf Avg.} \\ \cmidrule(lr){3-5} \cmidrule(lr){6-8}
 & \multicolumn{1}{c}{} & \multicolumn{1}{l}{MedQA} & \multicolumn{1}{l}{MedMCQA} & \multicolumn{1}{l}{MMLU$^*$} & \multicolumn{1}{l}{CMB} & \multicolumn{1}{l}{CMExam} & \multicolumn{1}{l}{CMMLU$^*$} &  \\ \midrule
 \multirow{3}{*}{Mistral-7B}  &\bf KaFT (Ours) & 59.54 & 49.87 & 68.47 & 38.11 & 37.35 & 40.73 & \bf 49.01 \\ 
 & -w. constant & 59.86 & 43.75 & 68.06 & 36.25 & 36.25 & 35.48 & 46.61 \\
 & -w. auto-adapt & 59.07 & 51.09 & 68.53 & 37.72 & 36.92 & 39.56 & 48.82 \\
\midrule
\multirow{3}{*}{Qwen1.5-7B}  & \bf KaFT (Ours) & 53.57 & 49.82 & 63.23 & 75.57 & 77.27 & 72.07 & \bf 65.25 \\
& -w. constant  & 52.71 & 50.20 & 61.60 & 74.30 & 76.15 & 70.16 & 64.19 \\
 & -w. auto-adapt & 52.55 & 50.06 & 63.04 & 75.18 & 77.04 & 70.73 & 64.77 \\
\midrule
\multirow{3}{*}{LLaMA3-8B}  & \bf KaFT (Ours) & 64.10 & 56.94 & 74.01 & 47.39 & 47.89 & 45.44 & \bf 55.96 \\
& -w. constant & 61.82 & 55.75 & 73.34 & 45.99 & 45.85 & 44.95 & 54.62 \\
 & -w. auto-adapt & 61.35 & 56.49 & 73.19 & 47.91 & 48.45 & 46.26 & 55.61 \\
\midrule
\multirow{3}{*}{LLaMA3-3B}  & \bf KaFT (Ours) & 54.52 & 50.51 & 64.93 & 40.19 & 39.93 & 39.70 & \bf 48.30 \\
& -w. constant & 54.99 & 50.08 & 62.94 & 38.00 & 39.17 & 38.13 & 47.22 \\
 & -w. auto-adapt & 53.57 & 50.30 & 63.65 & 39.15 & 39.55 & 38.17 & 47.40 \\
 \bottomrule
\end{tabular}
}
\caption{\textbf{Full results of Figure~\ref{fig:reward}, \textit{i.e.}, performance comparison (\%) between different reward strategies in KaFT}. The best average results are in \textbf{bold}.}
\label{tab:reward_all}
\end{table*}

\begin{table*}[t]
\centering
\setlength{\tabcolsep}{7pt}
\resizebox{\textwidth}{!}{%
\begin{tabular}{clrcllrccc}
\toprule
\multirow{2}{*}{\bf Backbone} & \multicolumn{1}{l}{\multirow{2}{*}{\bf Method}} & \multicolumn{3}{c}{\bf English Medical Benchmark} & \multicolumn{3}{c}{\bf Chinese Medical Benchmark} & \multirow{2}{*}{\bf Avg.} \\ \cmidrule(lr){3-5} \cmidrule(lr){6-8}
 & \multicolumn{1}{c}{} & \multicolumn{1}{l}{MedQA} & \multicolumn{1}{l}{MedMCQA} & \multicolumn{1}{l}{MMLU$^*$} & \multicolumn{1}{l}{CMB} & \multicolumn{1}{l}{CMExam} & \multicolumn{1}{l}{CMMLU$^*$} &  \\ \midrule
\multirow{9}{*}{Mistral-7B} & $\alpha$=0.1, $\beta$=0.1 & 57.03 & 48.10 & 68.84 & 37.67 & 37.79 & 40.54 & 48.33 \\
 & $\alpha$=0.1, $\beta$=0.5 & 59.54 & 49.87 & 68.47 & 38.11 & 37.35 & 40.73 &\bf 49.01 \\
 & $\alpha$=0.1, $\beta$=1.0 & 58.68 & 50.39 & 68.35 & 37.99 & 37.63 & 39.29 & 48.72 \\ \cmidrule{2-9}
 & $\alpha$=0.5, $\beta$=0.1 & 58.76 & 48.82 & 67.98 & 37.38 & 36.76 & 40.05 & 48.29 \\
 & $\alpha$=0.5, $\beta$=0.5 & 59.78 & 51.21 & 68.06 & 37.30 & 36.72 & 39.27 & 48.72 \\
 & $\alpha$=0.5, $\beta$=1.0 & 59.31 & 49.70 & 69.05 & 37.99 & 36.87 & 40.18 & 48.85 \\ \cmidrule{2-9}
 & $\alpha$=1.0, $\beta$=0.1 & 56.32 & 46.12 & 67.67 & 36.83 & 36.87 & 40.04 & 47.31 \\
 & $\alpha$=1.0, $\beta$=0.5 & 60.09 & 50.18 & 68.41 & 36.31 & 36.39 & 38.77 & 48.36 \\
 & $\alpha$=1.0, $\beta$=1.0 & 59.86 & 43.75 & 68.06 & 36.25 & 36.25 & 35.48 & 46.61 \\ \midrule
\multirow{9}{*}{Qwen1.5-7B} & $\alpha$=0.1, $\beta$=0.1 & 53.57 & 50.13 & 62.45 & 75.57 & 77.27 & 71.55 & 65.09 \\
 & $\alpha$=0.1, $\beta$=0.5 & 53.57 & 49.82 & 63.23 & 75.57 & 77.27 & 72.07 &\bf 65.25 \\
 & $\alpha$=0.1, $\beta$=1.0 & 52.47 & 50.18 & 62.53 & 75.25 & 76.89 & 71.46 & 64.80 \\  \cmidrule{2-9}
 & $\alpha$=0.5, $\beta$=0.1 & 52.95 & 49.65 & 61.96 & 75.14 & 76.99 & 71.94 & 64.77 \\
 & $\alpha$=0.5, $\beta$=0.5 & 53.97 & 50.04 & 61.24 & 75.09 & 76.69 & 71.30 & 64.72 \\
 & $\alpha$=0.5, $\beta$=1.0 & 53.42 & 50.27 & 60.82 & 74.59 & 76.19 & 70.90 & 64.37 \\  \cmidrule{2-9}
 & $\alpha$=1.0, $\beta$=0.1 & 52.87 & 50.59 & 62.35 & 75.17 & 76.48 & 70.45 & 64.65 \\
 & $\alpha$=1.0, $\beta$=0.5 & 51.53 & 50.08 & 61.73 & 74.86 & 76.13 & 70.49 & 64.14 \\
 & $\alpha$=1.0, $\beta$=1.0 & 52.71 & 50.20 & 61.60 & 74.30 & 76.15 & 70.16 & 64.19 \\
 \bottomrule
\end{tabular}
}
\caption{\textbf{Full results of Figure~\ref{fig:reward_analysis}, \textit{i.e.}, parameter analyses of $\alpha$ and $\beta$}. The best average results are in \textbf{bold}.}
\label{tab:parameter_analysis_all}
\end{table*}

\begin{table*}[t]
\centering
\setlength{\tabcolsep}{10pt}
\resizebox{\textwidth}{!}{%
\begin{tabular}{clccccccc}
\toprule
\multirow{2}{*}{\bf Backbone} & \multicolumn{1}{l}{\multirow{2}{*}{\bf Method}} & \multicolumn{6}{c}{\bf MMedBench} & \multirow{2}{*}{\bf Avg.} \\ \cmidrule(lr){3-8}
 & \multicolumn{1}{c}{} & Chinese & English & French & Japanese & Russian & Spanish &  \\ \midrule
\multirow{5}{*}{Mistral-7B} & Base & 35.76 & 51.06 & 41.74 & 26.88 & 48.05 & 49.27 & 42.13 \\
 & Vanilla SFT & 41.07 & 58.99 & 48.29 & 30.00 & 60.94 & 58.02 & 49.55 \\
 & No-conflict & 41.77 & 56.56 & 49.84 & 36.25 & 62.50 & 56.20 & 50.52 \\
 & Self-aligning & 41.04 & 54.91 & 46.11 & 32.50 & 61.33 & 55.87 & 48.63 \\
 & \bf KaFT (Ours) & 41.36 & 58.37 & 48.29 & 38.12 & 62.11 & 57.22 & \bf 50.91 \\ \midrule
\multirow{5}{*}{Qwen1.5-7B} & Base & 82.25 & 46.19 & 41.12 & 35.62 & 55.08 & 49.02 & 51.55 \\
 & Vanilla SFT & 79.16 & 46.82 & 45.48 & 36.25 & 61.72 & 49.12 & 53.09 \\
 & No-conflict & 83.07 & 50.90 & 48.60 & 44.38 & 57.81 & 52.88 & 56.27 \\
 & Self-aligning & 82.11 & 47.29 & 46.11 & 38.75 & 58.59 & 51.50 & 54.06 \\
 & \bf KaFT (Ours) & 82.81 & 51.61 & 47.66 & 40.62 & 63.67 & 52.63 & \bf 56.50 \\ \midrule
\multirow{5}{*}{LLaMA3-8B} & Base & 56.98 & 58.68 & 53.58 & 40.00 & 55.86 & 59.48 & 54.10 \\
 & Vanilla SFT & 58.20 & 61.74 & 57.01 & 42.38 & 63.67 & 61.93 & 57.49 \\
 & No-conflict & 59.40 & 60.72 & 58.57 & 46.25 & 62.50 & 61.42 & 58.14 \\
 & Self-aligning & 59.78 & 60.49 & 57.63 & 45.62 & 62.50 & 62.31 & 58.09 \\
 & \bf KaFT (Ours) & 60.19 & 63.24 & 58.88 & 46.25 & 63.67 & 62.00 & \bf 59.04 \\ \midrule
\multirow{5}{*}{LLaMA3-3B} & Base & 46.15 & 49.65 & 40.81 & 28.12 & 50.78 & 49.31 & 44.14 \\
 & Vanilla SFT & 47.14 & 53.10 & 40.81 & 33.75 & 51.95 & 51.79 & 46.42 \\
 & No-conflict & 48.63 & 53.57 & 42.06 & 35.00 & 51.17 & 51.17 & 46.93 \\
 & Self-aligning & 47.72 & 52.40 & 39.56 & 33.12 & 51.17 & 50.95 & 45.82 \\
 & \bf KaFT (Ours) & 48.22 & 53.42 & 46.11 & 33.12 & 51.95 & 52.81 & \bf 47.61 \\
 \bottomrule
\end{tabular}
}
\caption{\textbf{Full results of Figure~\ref{fig:mmedbench}, \textit{i.e.}, performance of MMedBench}~\cite{qiu2024towards}. In addition to LLaMA3-8B models, we also report the results of other LLMs. The best average results are in \textbf{bold}.}
\label{tab:mmedbench_all}
\end{table*}


\end{document}